%% file: iclr2027_conference.tex
\documentclass{article} 
\usepackage{iclr2027_conference,times}

\usepackage[textsize=scriptsize]{todonotes}
\newcommand{\jw}[1]{\todo[color=cyan!20,linecolor=cyan]{\textbf{JW:} #1}}
\usepackage{marginnote}
\usepackage{etoolbox}
\makeatletter
\patchcmd{\@todonotes@drawLineToRightMargin}{-| ([xshift=-0.2cm] inNote.west)-| (inNote.west);}{-- ([xshift=-0.3cm,yshift=-0.2cm] inNote.west |- inText) -- (inNote.north west);}{\typeout{JWPATCH OK}}{\typeout{JWPATCH FAILED}}
\makeatother
\newcommand{\jwfloat}[1]{{\let\marginpar\marginnote\jw{#1}}} 
\usepackage{xspace}
\newcommand{\ours}{PaTh\xspace}

\input{math_commands.tex}

\usepackage{hyperref}
\usepackage{url}
\usepackage{graphicx}
\usepackage{booktabs}
\usepackage{float}
\usepackage{multirow}
\usepackage{amsmath}
\usepackage{colortbl}
\usepackage{subcaption}
\usepackage{wrapfig}
\usepackage{tabularx}
\usepackage[normalem]{ulem}

\usepackage{algorithm}
\usepackage{algpseudocode}

\usepackage{xcolor}
\usepackage{comment}
\usepackage{pifont}
\definecolor{paththink}{HTML}{8B3FD9}  
\definecolor{dmred}{HTML}{E24B4A}
\definecolor{pathgreen}{HTML}{3B6D11}
\newcommand{\inlineicon}[1]{\raisebox{-0.15em}{\includegraphics[height=0.95em]{#1}}}
\newcommand{\thinkicon}{\inlineicon{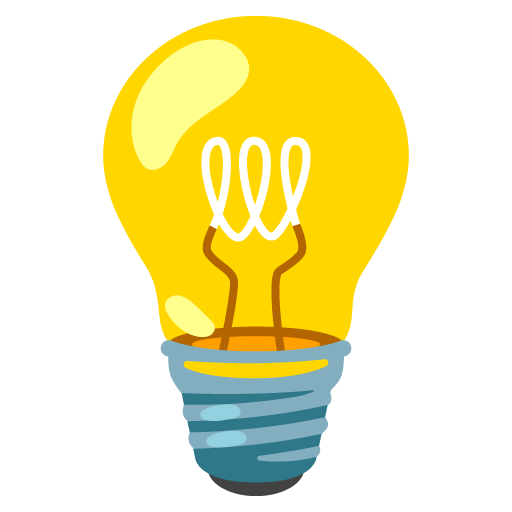}}
\newcommand{\painticon}{\inlineicon{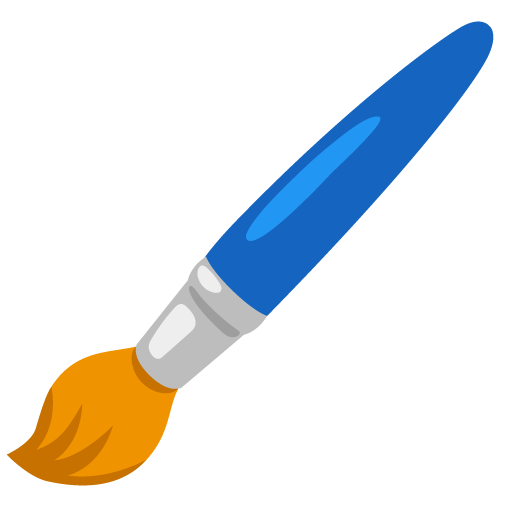}}
\newcommand{\cmark}{\textcolor{pathgreen}{\ding{51}}}
\newcommand{\xmark}{\textcolor{dmred}{\ding{55}}}

\definecolor{rowgray}{HTML}{D9D9D9}
\definecolor{rowyellow}{HTML}{FFF2CC}
\definecolor{rowred}{HTML}{FF9999}
\definecolor{groupbg}{gray}{0.93}  

\title{Think Before You Paint: Recursive Latent Reasoning for Diffusion Models}

\author{%
  Paweł Skierś$^{1,2}$ \quad
  Małgorzata Grzanka$^{1}$ \quad
  Wojciech Masarczyk$^{2}$ \\
  \textbf{Jan-Willem van de Meent$^{3}$ \quad
  Kamil Deja$^{1,2}$} \\[0.5em]
  $^{1}$Warsaw University of Technology \quad
  $^{2}$IDEAS Research Institute \quad
  $^{3}$University of Amsterdam \\[0.3em]
  \texttt{pawel.skiers.dokt@pw.edu.pl}
}

\iclrfinalcopy 
\begin{document}

\maketitle

\input{content}

\bibliography{iclr2027_conference}
\bibliographystyle{iclr2027_conference}

\appendix

\input{appendix}

\end{document}

%% file: math_commands.tex
\usepackage{amsmath,amsfonts,bm}

\def\eqref#1{equation~\ref{#1}}

\def\1{\bm{1}}

\DeclareMathAlphabet{\mathsfit}{\encodingdefault}{\sfdefault}{m}{sl}
\SetMathAlphabet{\mathsfit}{bold}{\encodingdefault}{\sfdefault}{bx}{n}



%% file: content.tex
\begin{abstract}

Diffusion models generate realistic images but often fail on visual reasoning tasks, such as filling in a Sudoku or drawing the path through a maze. When a discrete symbolic representation is available, recursive methods such as the Tiny Recursive Model (TRM) solve even hard instances of these puzzles. We ask how such reasoning can be carried over to pixels, where no symbolic representation is available. We propose Painter--Thinker (\ours): a small recursive network (the Thinker) reasons over a grid of learned tokens that encode the noisy image and the conditioning, refines a latent state within every denoising step, and steers a frozen diffusion model (the Painter) through ControlNet adapters. The Thinker is trained with the standard reconstruction loss alone, without symbolic targets, a solver or a verifier. \ours solves $92.5\%$ of hard MNIST Sudoku puzzles (prior best $75\%$) and $71.2\%$ of extreme ones (prior best $4.1\%$), with $10$M parameters against $82$M for a standard diffusion model. It also improves on mazes, Queens, and CLEVR scenes with specified spatial relations, and its advantage grows with problem size. Diagnostic experiments show that \ours recovers from injected mistakes that the diffusion model cannot repair, especially when many cells are wrong. Together, these results show that reasoning mechanisms developed for symbolic data can be integrated into pixel-space diffusion without symbolic supervision, opening a path toward generating data under increasingly complex constraints.

\end{abstract}

\section{Introduction}

Diffusion models (DM) are highly effective at modeling continuous data distributions~\citep{ho2020ddpm}, and under the right representation they can satisfy hard constraints: discrete DMs solve Sudoku when the puzzle is given symbolically \citep{ye2025beyondar}, and equivariant diffusion generates valid molecules \citep{hoogeboom2022edm}. In both cases the representation does much of the work, since the rules apply to units the model already manipulates: cells, atoms, or bonds, where a violation is fixed by changing one of them. Success is far less consistent once the same rules must hold over a continuous signal that merely depicts those units: denoising models produce traffic scenes in which vehicles collide or spawn inside buildings \citep{chauhan2025vfsi}, velocity fields that are not divergence-free \citep{pcdm2026}, and puzzles whose cells are individually legible but jointly invalid.

In this work, we study this failure and ask whether \emph{diffusion models can learn to follow logical rules directly from pixels?} 
To separate constraint satisfaction from representational difficulty, we turn to visual puzzles such as a Sudoku rendered as a grid of MNIST digits~\citep{wewer2025srm}, a maze drawn as an image, an $N$-queens board~\citep{zhou2026probing}, and compositional CLEVR scene generation~\citep{johnson2017clevr}.
These setups are visually simple, and the rules are exactly verifiable. Vanilla diffusion models however, solve only
$8.3\%$ of hard MNIST Sudoku instances 
despite producing sharp and diverse MNIST digits, and simply adding model capacity does not improve reasoning performance.

\begin{figure}[t]
    \centering
    \vspace{-1em}
    \includegraphics[width=\linewidth]{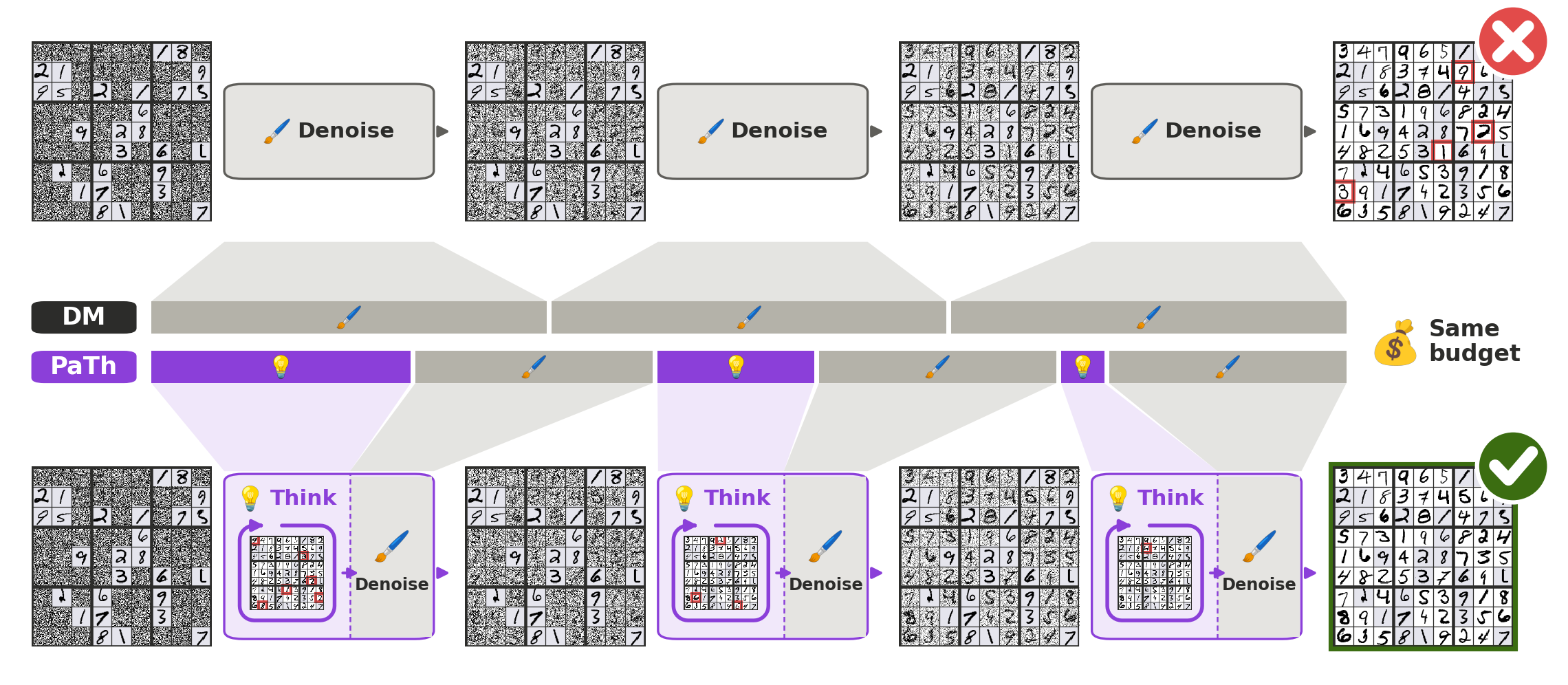}      
    \caption{\textbf{Diffusion models are not limited by scale but by where they spend compute.}
  \emph{Top:} a standard diffusion model (DM) spends its entire inference budget on
  denoising (\painticon) and commits to errors early, leaving an invalid sudoku (\xmark).
  \emph{Bottom:} PaTh (Painter--Thinker) pairs a smaller denoiser, the \emph{Painter},
  with a recursive \textcolor{paththink}{\emph{Thinker}} (\thinkicon, loop) that refines
  a draft of the solution before each denoising step, with most thinking at high noise,
  and solves the puzzle (\cmark). Both models have the same total inference budget. On MNIST Sudoku \textsc{extreme}, PaTh solves \textbf{71.2\%} of puzzles versus \textbf{2.9\%} for the vanilla DM
  (Sec.~\ref{sec:experiments}). \vspace{-1em}}
  \label{fig:teaser}
\end{figure}

Resolving such rules requires serial computation~\citep{merrill2024expressive, li2024serial}. Recursive reasoners supply this computation by repeatedly refining a latent state: the Tiny Recursive Model (TRM) and the Hierarchical Reasoning Model (HRM) solve hard symbolic puzzles this way~\citep{jolicoeurmartineau2025trm, wang2025hrm}, and recent work shows that the same recurrence, carried along a flow over tokens, reaches $97.9\%$ on Sudoku-Extreme~\citep{suleymanzade2026loopedflows}. Carrying this over to pixels is not trivial, because in pixel space the reasoner is never told which symbols the image depicts. 
Simply replacing the denoiser's architecture with a recursive reasoner yields only a marginal improvement (Tab.~\ref{tab:sudoku_results}), as it asks a single network to both reason about which sample satisfies the rule and render that sample in pixels.
We therefore separate the two functions. In what we call a Painter-Thinker architecture (\ours), a diffusion model (the Painter) learns the data distribution and is then held fixed, while a small recursive reasoner (the Thinker) reasons over a grid of learned tokens that encode the noisy sample and the conditioning, refines a latent state within each denoising step, and steers the Painter through ControlNet adapters. As these refinement steps are decoupled from image generation, the Thinker can explore and revise candidate solutions before anything is committed to the sample. Both modules are trained with a reconstruction loss alone, without a solver, a verifier, or symbolic targets.

In a series of experiments across diverse benchmarks, we show that adding a Thinker buys more than adding parameters. \ours{} improves on the previous best of \citet{kang2026ipr} by over $67$~p.p.\ on MNIST Sudoku \textsc{extreme} and surpasses diffusion models with $700\times$ its capacity on AMAZE~\citep{zhou2026probing}. The trend continues as tasks grow harder: on complex 3D reasoning, the Painter must grow to cover a richer data distribution while the Thinker that steers it stays comparatively small, with a $20$M-parameter Thinker directing a $100$M-parameter Painter.
We perform a series of mechanistic diagnostics to establish that adding the Thinker improves reasoning. Corrupting a partially generated sample shows that the diffusion model repairs violations already visible in the sample, but fails to indicate what to trust when too much of the sample is wrong, whereas \ours recovers across both regimes. Probing the Thinker’s latent state shows candidates that violate more constraints than their predecessors before a solution is committed to the image.
Our contributions are as follows.
\begin{itemize}
    \item We show that coupling a TRM-style recursive reasoner to a frozen diffusion model via ControlNet adapters effectively adds capacity for visual reasoning.
    \item We demonstrate consistent improvements over standard diffusion models, diffusion methods that adapt the sampling process, and large image-editing models on four families of rule-governed data, with margins that widen as instances grow harder.
    \item We diagnose the mechanism behind these gains: \ours's latent state supports iterative refinement before committing to the output, which allows it to recover from mistakes. 
\end{itemize}

\section[Related work]{Related work\protect}
\paragraph{Rule-governed generation with diffusion.}
Several methods make diffusion models satisfy rules in pixel space by changing how the trajectory is sampled. Models trained in pixel space can approximate solutions to hard geometric problems \citep{goren2026geometric}, and such interventions become necessary on rule-governed data.    
One option is to restructure the trajectory itself: SRM \citep{wewer2025srm} decomposes
an image into variables with individual noise schedules and denoises them in an
order predicted from the model's own uncertainty. 
Other methods supply the reasoning from outside the model by optimising a learned energy
landscape at inference \citep{du2024ired}, importing it from a pretrained
VLM \citep{mi2025thinkdiff}, or delegating a plan to a
multimodal LLM before generation \citep{yang2024rpg}. Benchmarks for such problems have been assembled by
\cite{wewer2025srm, zhou2026probing}.
While SRM shares our diagnosis that compute is misallocated along the
trajectory, it redistributes steps that still bear commitments. 
Relatedly, \citet{wang2026demystifying} argue that reasoning in video models emerges along
the denoising steps rather than across frames, with early steps exploring
multiple candidates. Our results in \ref{sec:diagnosis} are consistent with this observation.

\paragraph{Inference-time compute and guidance.}
A second way to improve a sample is to spend more compute searching for it at sampling time: search over initial noise against a verifier
\citep{ma2025inferencescaling}, sequential Monte Carlo steering
\citep{singhal2025fksteering}, evolutionary search \citep{he2025evosearch}, tree
search over denoising trajectories \citep{yoon2025mctd}, and resampling schemes
that revisit their parts \citep{lugmayr2022repaint, xu2023restart}. Closest to our setting, IPR \citep{kang2026ipr} re-noises and regenerates
subsets of a completed sample without a verifier.
A parallel line steers sampling towards a constraint through gradients of a
classifier, a differentiable objective, or a composed energy
\citep{dhariwal2021classifierguidance, ho2022cfg, chung2023dps, yu2023freedom,
bansal2023universal, liu2022composable}. Both families revise in data space,
requiring either an external verifier or a differentiable constraint. \ours has neither, and revises before anything
enters the sample. 

\paragraph{Discrete diffusion for reasoning.}
Diffusion does reason well when the data is symbolic. Discrete DMs
\citep{austin2021d3pm, sahoo2024mdlm} and the masked language models built on
them \citep{nie2025llada, ye2024dot, ye2025beyondar} apply denoising to tokens,
where reasoning has been studied directly. The discrete setting affords
what continuous one does not: a masked token is explicitly undecided,
so a partial state does not have to be a complete sample. In continuous diffusion, every
$\mathbf{x}_t$ is a full sample with no way to mark part of it as pending, which
is the gap we fill.

\paragraph{Intermediate computation and recursive reasoners.}
Decoupling the state used for intermediate computation and the the output comes from language models and recursive reasoners. In particular, chain-of-thought extends the serial depth available to a fixed-depth model, and this is what
allows it to solve inherently serial problems \citep{merrill2024expressive,
li2024serial}; the intermediate tokens need not even be meaningful for the
benefit to appear \citep{pfau2024dotbydot}, and the same computation can be
carried out in a continuous latent space instead \citep{hao2025coconut}. Recursive
reasoners realise this through weight-tied depth
\citep{dehghani2019universal, bai2019deq}, with HRM \citep{wang2025hrm} and TRM
\citep{jolicoeurmartineau2025trm} reaching strong results on puzzles at a few
million parameters. Concurrent work couples such recurrence with flows over tokenised
puzzles, through self-conditioning \citep{helbling2026frm} or a latent state
carried along the flow \citep{suleymanzade2026loopedflows}, and finds that a
flow without recurrence falls well short, consistent with our diagnosis. \ours differs from all of
these in targeting exactly checkable rules in pixel space.

\section{Method}
\label{sec:method}
\subsection{Preliminaries}
\textbf{A diffusion model} generates a sample by traversing a sequence of states
$\mathbf{x}_T, \dots, \mathbf{x}_0$, at each step predicting and removing noise
so that $\mathbf{x}_{t-1}$ is a less noisy version of the same sample
\citep{ho2020ddpm}. Every state on this trajectory is a partially denoised
sample. We use the $\mathbf{x}_0$-parameterisation throughout: the network
predicts the clean sample directly,
$\hat{\mathbf{x}}_0 = f_\theta(\mathbf{x}_t, t, \mathbf{c})$, conditioned on an
embedded input $\mathbf{c}$, and is trained with a reconstruction loss alone,
\begin{equation}
    \mathcal{L}(\theta) = \mathbb{E}_{\mathbf{x}_0, t, \epsilon}
    \big\lVert \mathbf{x}_0 - f_\theta(\mathbf{x}_t, t, \mathbf{c}) \big\rVert^2 .
    \label{eq:recon}
\end{equation}

\textbf{A recursive reasoner} maintains a current answer $\mathbf{y}$ and a latent
reasoning state $\mathbf{z}$, and refines both against a fixed embedded input
$\mathbf{c}$. Computation is organised in three
nested levels. Outermost are $N_{\text{sup}}$ deep supervision steps; each carries
a detached $(\mathbf{y}, \mathbf{z})$ into the next. Within a supervision step the
answer is updated $H$ times, of which the first $H-1$ run without gradients. Each
answer update is preceded by $L$ updates of the latent,
using the same network for both; the two roles are distinguished only by whether
$\mathbf{c}$ is supplied. The effective depth of a single forward pass is therefore
$N_{\text{sup}} \cdot H \cdot (L+1)$ applications of $\mathrm{net}$, at the parameter
cost of one. Only $\mathbf{y}$ is read out through an output head. The latent
$\mathbf{z}$ is never decoded and does not correspond to a solution at all
\citep[Fig.~6]{jolicoeurmartineau2025trm}, so it is free to hold whatever supports
the next update of $\mathbf{y}$.

\subsection{Painter and Thinker}

\begin{figure}[t]
    \centering
    \includegraphics[width=\linewidth]{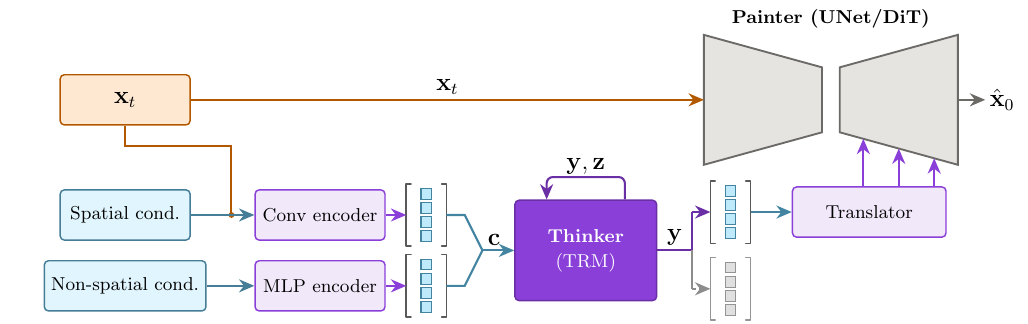}
    \caption{\ours{} architecture. Spatial conditioning is concatenated with
$\mathbf{x}_t$ and encoded convolutionally; non-spatial conditioning is embedded
by an MLP. Together they form the Thinker's input $\mathbf{c}$, over which the
Thinker recurses on $(\mathbf{y}, \mathbf{z})$ within a single denoising step.
Only the spatial tokens of $\mathbf{y}$ are decoded and passed to the Painter
through ControlNet adapters; the non-spatial tokens (gray squares) shape the
recursion but are not read out. The Painter is trained separately and held
fixed.\vspace{-1.5em}}
    \label{fig:architecture}
\end{figure}

\ours (visualized in Fig.~\ref{fig:architecture}) places a recursive reasoner inside a single denoising step. 
Given $\mathbf{x}_t$ and the conditioning, the Thinker refines a latent candidate over
$N_{\text{sup}}$ supervision steps and returns a signal which conditions the Painter's
prediction of $\mathbf{x}_{t-1}$. 

\paragraph{Encoding.} Spatial conditioning, such as the incomplete puzzle, is
concatenated channel-wise with $\mathbf{x}_t$, passed through 
a small convolutional network and mean-pooled to a fixed number of tokens. Non-spatial
conditioning, such as the object and relation vectors of CLEVR, is embedded by
an MLP and appended. The result is the Thinker's input $\mathbf{c}$, with
$\mathbf{y}$ and $\mathbf{z}$ initialised as in TRM.

\paragraph{Decoding and coupling.} After the final supervision step, the spatial
tokens of $\mathbf{y}$ are upsampled by a small transposed convolutional network
and passed through a convolutional head; any non-spatial tokens inform the
Thinker but are discarded here. The head's outputs enter the Painter through
ControlNet adapters \citep{zhang2023controlnet}, which add them to each skip
connection of the UNet, after which the Painter performs its usual denoising
step. We also train a halting head to predict the expected improvement from
an additional supervision step, following the adaptive-computation formulation
of TRM. We use it to reduce the training time.


\paragraph{Two trajectories.} \ours thus carries two sequences of states, and
they differ in kind. The states $\mathbf{x}_t$ are partial samples, each
committed to the output. The states $\mathbf{z}$ are not: they are refined
within a step, read out only through $\mathbf{y}$, and reset before the next 
step begins. Nothing requires an intermediate $\mathbf{z}$ to correspond to a
plausible image, or even to a better candidate than its predecessor, which is
what allows the Thinker to abandon a partial solution. 
Counting layers, the reasoning performed within one denoising step has an
effective depth of $n_{\text{layers}}(L+1)HN_{\text{sup}}$, or $672$ in our configuration, against the
single-network evaluation a diffusion model performs at that step.

\paragraph{Efficient inference.} Such depth is not needed at every step. As sampling
proceeds, $\mathbf{x}_t$ becomes cleaner and more of the solution is already
visible, leaving less to be reasoned about~\citep{pan2024tstitch, wewer2025srm}.We exploit this observation to reduce the compute needed during training (see Figures~\ref{fig:budget_allocation_ablation},~\ref{fig:reuse_steering}.
Nor does the Thinker's state need to be rebuilt at each step: it retains what
has already been worked out, and the conditioning remains fixed throughout
sampling. We therefore define an efficient variant of \ours that carries the
Thinker's state across consecutive denoising steps, concentrates recursion in
the early part of the trajectory, and then stops calling the Thinker halfway
through, reusing its last output for the remaining steps. We evaluate this
variant in Sec.~\ref{sec:compute}.

\paragraph{Design choices.} For the Painter, we use a UNet denoiser, but we show that \ours adapts to DiT in App.~\ref{app:adoption_to_dit}. The Thinker is a recursive reasoner with self-attention layers. We couple the two models via ControlNet adapters, but test concatenation of Thinker's output to $\mathbf{x}_t$ as an alternative in App.~\ref{app:concat_vs_controlnet}. We train the Painter and Thinker separately, as joint training takes longer and produces less diverse samples (see App.~\ref{app:joint_vs_separate}). We train \ours with $\mathbf{x}_0$-parameterisation as opposed to $\mathbf{eps}$, following \citep{sangare2026improving}, which finds that it is preferable when using ControlNets. In App.~\ref{app:stability}, we show that \ours is stable across different numbers of tokens the Thinker processes. Parameter counts and training budgets are in Tab.~\ref{tab:param_counts}, while hyperparameters in App.~\ref{app:implementation_details}.

\section{Experiments}\label{sec:experiments}

\paragraph{Datasets.} We use three datasets (details and examples in App.~\ref{app:datasets}).
\textbf{MNIST Sudoku}~\citep{wewer2025srm}: filling in an incomplete Sudoku rendered with MNIST digits. In \textsc{hard}, 0--26 random clues usually allow many solutions; in \textsc{extreme}~\citep{jolicoeurmartineau2025trm}, about 18 clues fix a unique solution that requires multi-step deduction.
\textbf{AMAZE}~\citep{zhou2026probing}: drawing maze paths and solving Queens boards of increasing size.
\textbf{CLEVR}~\citep{johnson2017clevr}: scenes of 3--10 objects generated from a sparse set of pairwise spatial relations. The model must compose these relations transitively to place every object.

\subsection{Main results}
\begin{wraptable}{r}{0.35\linewidth}
    \centering
    \vspace{-1em}
    \small
    \begin{tabular}{l c c}
        \toprule
        Model & Hard & Extreme \\
        \midrule
        DM & 0.083& 0.029\\
        TRM & - & 0.083 \\
        SRM & 0.516 & 0.032\\
        IPR & 0.750 & 0.041\\
        \ours & \textbf{0.925} & \textbf{0.712}\\
        \bottomrule
    \end{tabular}
    \caption{Puzzle accuracy on MNIST Sudoku. While \textsc{hard} puzzles usually admit many solutions \textsc{extreme} has exactly one. \ours significantly outperforms other methods on both setups.\vspace{-1            em}}
    \label{tab:sudoku_results}
\end{wraptable}
\paragraph{MNIST Sudoku.} Tab.~\ref{tab:sudoku_results}  reports results on both
variants of the MNIST Sudoku dataset (hard and extreme). \ours attains $92.5\%$ puzzle accuracy on \textsc{hard} and $71.2\%$ on \textsc{extreme}, exceeding every baseline by a wide margin. The gap is largest on \textsc{extreme}, where each puzzle admits a single solution and one misplaced digit renders the grid invalid. 
Notably, almost all of the capacity in \ours is devoted to reasoning rather than
to rendering. The Painter has $1.5$M parameters and the Thinker $8$M, against
$81$M for the DM baseline (Tab.~\ref{tab:param_counts}). 
Rendering MNIST
digits is easy, and the difficulty resides in the rules, so separating
the two permits capacity to be allocated where it is needed. 
SRM and IPR presuppose a
decomposition of the image into variables aligned with the $81$ Sudoku cells, which encode the structure of
the puzzle. The Thinker requires no such alignment and performs comparably
across patch resolutions that bear no relation to the cell grid
(App.~\ref{app:stability}).

\begin{figure}[b]
    \centering
    \includegraphics[width=1.0\linewidth]{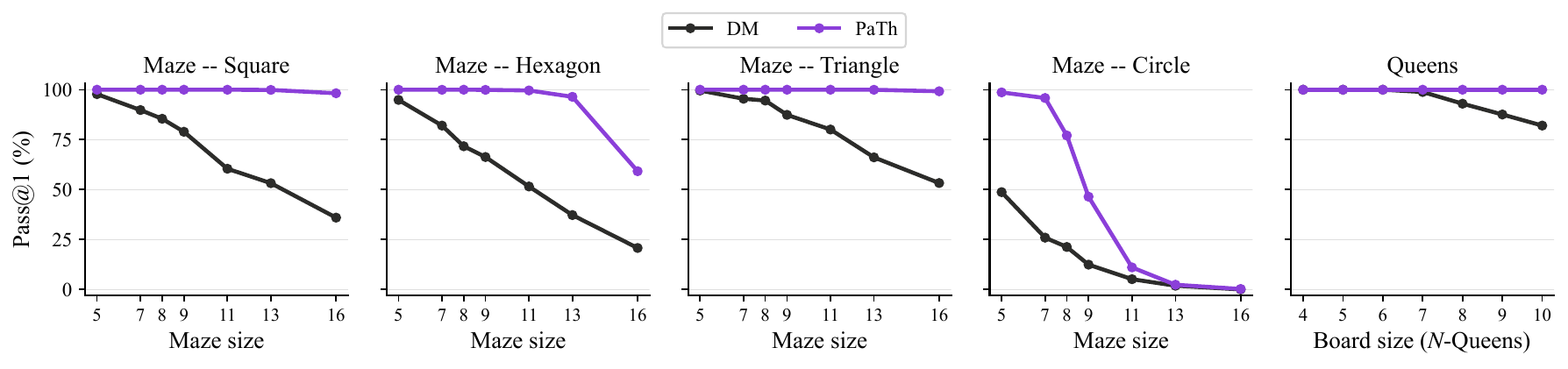}
    \caption{On Amaze benchmark, \ours has an increasing advantage over standard DM as the complexity of the puzzles increases - x-axis: puzzle size, y-axis: pass@1; for both datasets \vspace{-1em}}
    \label{fig:amaze_dm_vs_ours}
\end{figure}

\paragraph{AMAZE benchmark.} Tab.~\ref{tab:results_amaze} compares \ours with
proprietary and open-source image editing models and a DM trained identically on the same data. \ours solves $75.1\%$ of mazes
exactly (\textsc{exact}@$1$), against $43.4\%$ for the strongest editing model,
a fine-tuned 14B Bagel, and $19.0\%$ for the DM. On Queens, it solves every
board, against $90.9\%$ for the DM and $13.7\%$ for fine-tuned Bagel. Fine-tuned Bagel attains slightly higher \textsc{pass} and \textsc{coverage} on
mazes. While it typically draws the correct path in full, it leaves faint traces of other paths, which
keep most of its generations from being exactly correct
(App.~\ref{app:bagel_traces}). This is consistent with our account: a model
whose only state is the image must record every path it tries in the image,
whereas \ours explores in its latent state and leaves the output clean.

\begin{table*}[t!]\footnotesize
\renewcommand{\arraystretch}{1}
\setlength{\tabcolsep}{2pt}
\resizebox{\textwidth}{!}{%
\begin{tabular}{l@{\hspace{1pt}} c@{\hspace{5pt}} cccccc @{\hspace{20pt}} cccccc}
\toprule
\multirow{2}{*}{\textbf{Model}} & \multirow{2}{*}{\textbf{Params}} & \multicolumn{6}{c}{\textbf{Continuous (Maze) Task}} & \multicolumn{6}{c}{\textbf{Discrete (Queen) Task}} \\
\cmidrule(lr{20pt}){3-8} \cmidrule(lr){9-14}
& & \textbf{Violation}$\downarrow$ & \textbf{Coverage}$\uparrow$ & \textbf{Pass@1}$\uparrow$ & \textbf{Pass@5}$\uparrow$ & \textbf{Exact@1}$\uparrow$ & \textbf{Exact@5}$\uparrow$& \textbf{Violation}$\downarrow$ & \textbf{Coverage}$\uparrow$ & \textbf{Pass@1}$\uparrow$ & \textbf{Pass@5}$\uparrow$ & \textbf{Exact@1}$\uparrow$ & \textbf{Exact@5}$\uparrow$ \\
\midrule
\rowcolor{groupbg}
\multicolumn{14}{c}{\textit{proprietary models}} \\
\midrule
GPT-image-1 & N/A & 62.88 & 58.97 & 5.40 & 6.06 & -- & -- & 62.91 & 37.09 & 0.00 & 2.28 & -- & -- \\
NanoBanana-Pro & N/A & 47.76 & 64.21 & 4.82 & 9.28 & -- & -- & 32.56 & 67.43 & 30.35 & 35.58 & -- & -- \\
Seedream-4.5 & N/A & 16.90 & 25.67 & 2.14 & 3.21 & -- & -- & 76.86 & 23.14 & 2.86 & 2.86 & -- & -- \\
\midrule
\rowcolor{groupbg}
\multicolumn{14}{c}{\textit{open-source models (w/o chain-of-thought reasoning)}} \\
\midrule
Flux-Kontext-Dev & 12B & 23.84 & 30.24 & 0.36 & 3.57 & 0.00 & 0.00 & 78.63 & 21.37 & 0.92 & 2.34 & 0.00 & 0.00 \\
Qwen-Image-Edit & 20B & 19.37 & 28.51 & 1.43 & 2.14 & 0.18 & 0.36 & 69.52 & 30.47 & 2.86 & 4.00 & 0.16 & 0.24 \\
Bagel & 14B & 28.91 & 27.15 & 0.00 & 1.00 & 0.00 & 0.00 & 61.57 & 38.43 & 0.00 & 0.00 & 0.00 & 0.00 \\
Janus-Pro & 7B & \textbf{5.41} & 1.85 & 0.00 & 0.00 & 0.00 & 0.00 & 84.24 & 15.76 & 0.00 & 0.57 & 0.00 & 0.00 \\
Bagel (FT) & 14B & 12.82 & \textbf{99.99} & \textbf{87.17} & \textbf{87.21} & \uline{43.39} & \uline{52.75} & 21.18 & \uline{99.68} & 78.71 & 78.67 & 13.71 & 37.71 \\
Janus-Pro (FT) & 7B & 41.02 & 48.67 & 18.03 & 17.85 & 0.04 & 0.36 & 74.73 & 13.23 & 0.72 & 1.04 & 0.00 & 0.29 \\
\midrule
\rowcolor{groupbg}
\multicolumn{14}{c}{\textit{ours (diffusion models)}} \\
\midrule
DM & 62.7M & 23.15 & 77.32 & 57.78 & 57.73 & 18.96 & 22.75 & \uline{2.84} & 97.02 & \uline{94.50} & \uline{94.71} & \uline{90.86} & \uline{96.00} \\
\ours (ours) & 10.3M & \uline{8.64} & \uline{92.30} & \uline{85.30} & \uline{85.28} & \textbf{75.14} & \textbf{79.25} & \textbf{0.00} & \textbf{100.00} & \textbf{100.00} & \textbf{100.00} & \textbf{100.00} & \textbf{100.00} \\
\bottomrule
\end{tabular}
}
\caption{Results on the AMAZE benchmark's Maze and Queens tasks. \ours is able to outperform both standard DMs trained on the same data and the image editing models evaluated on the benchmark, including their fine-tuned (FT) versions and proprietary systems.}
\label{tab:results_amaze}
\vspace{-1em}
\end{table*}

Fig.~\ref{fig:amaze_dm_vs_ours} resolves the comparison against that diffusion baseline by instance size. On hexagon mazes, both models solve the smallest instances, but the baseline falls below $50\%$ after size $11$, whereas \ours remains above $50\%$ even at the biggest size. The pattern repeats on square and triangle mazes, and on Queens, \ours is unaffected across the full range of board sizes while the baseline degrades steadily. Larger instances require longer chains of dependent inferences, which the denoising trajectory has no means of accumulating, so the widening gap is the behaviour our account predicts. A related failure is reported for editing models on AMAZE, which produce valid path fragments near the start and goal but cannot join them at larger scales \citep{zhou2026probing}. Circular mazes are the exception, as both models degrade sharply and \ours retains an advantage only up to size $11$.

\begin{wrapfigure}{r}{0.41\linewidth}
\vspace{-1.5em}
    \begin{minipage}[b]{\linewidth}
    \centering
    \small
    \setlength{\tabcolsep}{3pt}
    \captionof{table}{Constrained scene composition on CLEVR (\%). \ours substantially improves relational consistency while maintaining object and attribute metrics.}
    \label{tab:results_clevr}
    \begin{tabular}{l c c c c}
        \toprule
        Method & Recall & Prec. & Attr. & Spatial \\
        \midrule
        DM (101M)    & 96.5 & \textbf{97.5} & 54.6 & 62.3 \\
        DM (124M) & 96.5 & \textbf{97.5} & 56.1 & 64.3 \\
        DM (236M) & \textbf{96.9} & \textbf{97.5} & \textbf{56.3} & 68.8 \\
        \ours & 95.9 & 97.5 & 53.8 & \textbf{89.4} \\
        \bottomrule
    \end{tabular}
    \end{minipage}
\vspace{-1em}
\end{wrapfigure}
\begin{figure}[t]
    \centering
    \begin{minipage}[t]{0.43\linewidth}
        \centering
        {\footnotesize\hspace*{24pt}\textbf{CLEVR}}\\[2pt]
        \includegraphics[height=3.9cm,keepaspectratio]{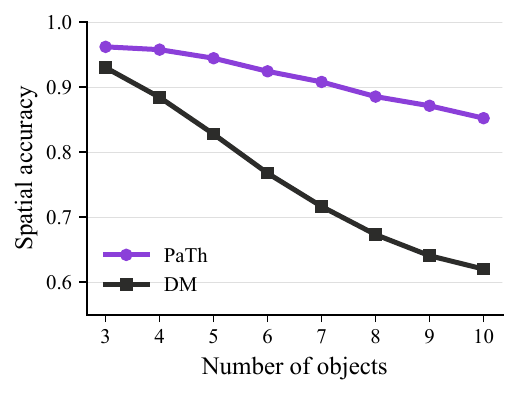}
        \caption{Spatial accuracy on CLEVR by scene size. The DM (238M) falls rapidly, while \ours retains high accuracy.}
        \label{fig:clevr_gains}
    \end{minipage}
    \hfill
    \begin{minipage}[t]{0.54\linewidth}
        \centering
        {\footnotesize\hspace*{18pt}\textbf{Sudoku}}\\[2pt]
        \includegraphics[height=3.9cm,keepaspectratio]{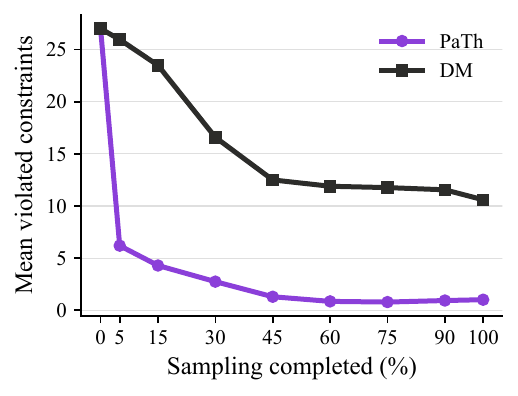}
        \caption{Sudoku constraints violated by the predicted clean
        sample $\hat{\mathbf{x}}_0$ during sampling. Both start near $27$;
        DM plateaus at $10.6$, while \ours falls below $1.5$.}
        \label{fig:constraint_violation_trajectory}
    \end{minipage}
    \vspace{-1.5em}
\end{figure}

\paragraph{Constrained scene composition.} We finally extend \ours beyond puzzles. To that end, we use CLEVR as a proof of concept in a domain where rendering
is genuinely demanding, so the Painter is a LDM of $101.9$M
parameters paired with a $19.7$M Thinker (Tab.~\ref{tab:param_counts}). 
As presented in Tab.~\ref{tab:results_clevr}, both standard DM and \ours are able to generate high-quality samples as denoted by high Precision and Recall, with comparable correspondence to their attributes. However, our approach outperforms the baseline in spatial accuracy, which rises from $62.3\%$ to $89.4\%$. 
This advantage persists even with the larger $238$M DM, which reaches $68.8\%$, showing that the gain is not simply a consequence of a bigger parameter count.
The Thinker improves what the model gets right about the relations between objects and leaves the objects themselves untouched, which is the division the method anticipates: only the relational constraint demands intermediate computation. The advantage also grows with scene complexity (Fig.~\ref{fig:clevr_gains}), from a near tie on three-object scenes to $85.2\%$ against $62.0\%$ at ten objects, mirroring the behaviour observed on AMAZE.

\subsection{Diagnosing the mechanism}
\label{sec:diagnosis}
The results presented in the previous section establish that \ours outperforms DMs in complex rule-based scenarios. We now isolate the source of this difference.
\paragraph{Constraint violations during sampling.} We first ask whether the standard diffusion model learns the underlying rules at all, using the sudoku dataset. To measure this, we classify each digit cell of the predicted clean sample $\hat{\mathbf{x}}_0$ at every denoising step and count the constraints it violates, averaged over $384$ samples (Fig.~\ref{fig:constraint_violation_trajectory}). Both trajectories begin near $27$ violations, and both decrease, so neither model is indifferent to the rule. They diverge in how far they get: the diffusion model plateaus at around $10.6$ violations, never approaching a valid grid, whereas \ours reaches fewer than $1.5$. \ours moreover achieves most of this reduction within the first $5\%$ of sampling, falling to $6.3$ violations before the image has taken definite form, a concentration we return to in Sec.~\ref{sec:compute}.


\paragraph{Recovering from injected mistakes.} To compare how the two models handle errors, we corrupt a valid solution, re-noise it to a chosen point of the trajectory, and let each model denoise it. On Sudoku we change digits of an \textsc{extreme} solution, and count a sample as recovered if it returns to that solution, which is unique. On CLEVR we swap the positions of pairs of objects, and count a sample as recovered if every object is placed correctly.


\begin{wrapfigure}{r}{0.45\linewidth}
\vspace{-1em}
    \centering
    \includegraphics[width=\linewidth]{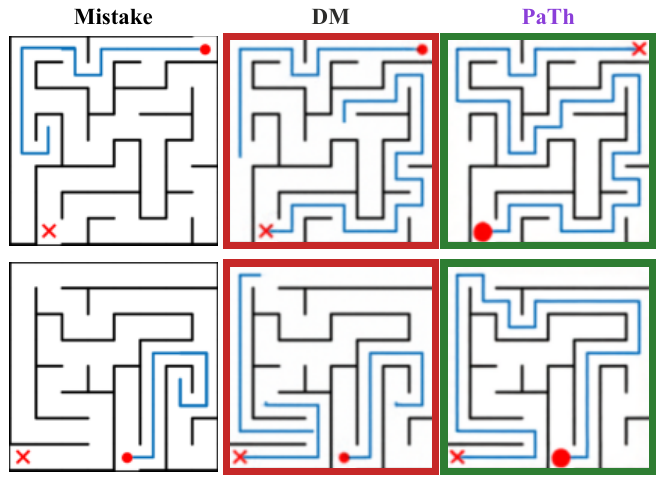}
    \caption{Qualitative comparison on mazes. When the path is directed into a dead-end, only \ours is able to correctly revise it.
    }
    \label{fig:qualitative_global_vs_local}
    \vspace{-2em}
\end{wrapfigure}

In Fig.~\ref{fig:fix_mistakes_recovery_rates}, we present the results of this comparison. Past roughly half the trajectory, both models deteriorate towards zero. Before that point, the diffusion model is effective only within a narrow band -- at 50\% of sampling on Sudoku and at the very start on CLEVR -- recovering over 70\% of single Sudoku mistakes at its peak and 27\% swaps at 5 object scenes, but failing significantly at other stages of sampling. It also degrades sharply as mistakes accumulate (Sudoku) or scene complexity increases (CLEVR). In contrast, \ours shows neither pattern, maintaining high recovery rates even when a vast majority of cells are wrong early in the process. It declines mainly with the timestep of corruption, and is even eventually overtaken by the diffusion model at the 70\% mark on the Sudoku dataset.



Both models are limited by how much of the trajectory remains, since errors introduced late cannot be substantially changed. We hypothesise that the diffusion model is limited further. Having no state apart from the sample, it can correct a mistake only if the mistake is \emph{directly available}: the sample must show both what is wrong and what should replace it. Early in sampling, the context needed for this is not yet formed; late in sampling, the relevant content is already committed. This also explains why recovery degrades with heavier Sudoku corruption: when too many digits are wrong, the sample provides less reliable context for distinguishing correct from incorrect entries. Likewise, more complex CLEVR scenes require longer chains of spatial relations to determine where an object belongs, making the required correction less directly available in the sample.


\begin{figure}[t]
    \centering
    \begin{minipage}[b]{0.56\linewidth}
        \centering
        {\footnotesize\hspace*{10pt}\textbf{Sudoku}}\\[2pt]
        \includegraphics[height=3.5cm]{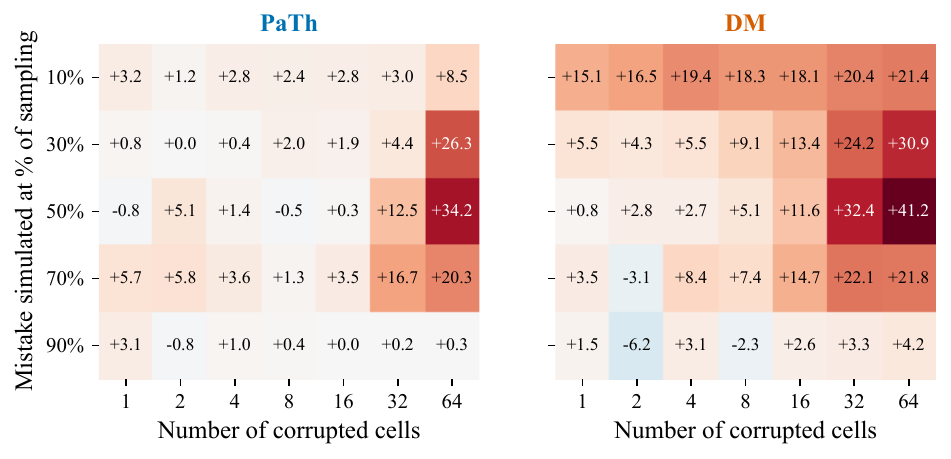}
    \end{minipage}
    \hfill
    \begin{minipage}[b]{0.425\linewidth}
        \centering
        {\footnotesize\hspace*{12pt}\textbf{Mazes}}\\[2pt]
        \includegraphics[height=3.5cm]{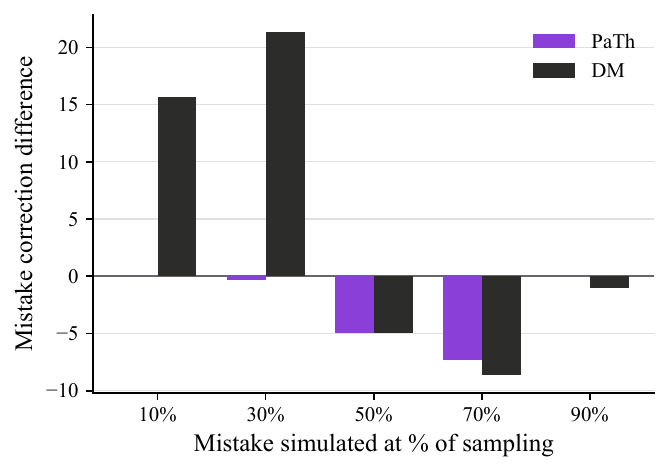}
    \end{minipage}
    \caption{Difference in retention rate between indirectly and directly available mistakes on \textsc{extreme} MNIST Sudoku (left) and Mazes (right). The diffusion model retains the indirect mistakes far more often early in sampling, while \ours treats the two types of errors comparably.}
    \label{fig:keep_mistake_local_vs_global}
    \vspace{-1em}
\end{figure}

To test this account, we vary the direct availability of mistakes while holding the timestep fixed on Sudoku and mazes. For Sudoku, directly available mistakes duplicate a digit already present in their row, column, or block, making the conflict explicit from the conditioning alone; indirectly available mistakes produce no such duplication but remain wrong because the \textsc{extreme} puzzle has a unique solution. For mazes, directly available mistakes remove part of the correct trajectory, whereas indirect mistakes divert the trajectory into a dead end, requiring the model to identify the wrong turn. Thus, only indirect mistakes must be worked out rather than directly identified, predicting a large gap between the two for DM but a much smaller gap for \ours.
Figure~\ref{fig:keep_mistake_local_vs_global} confirms this prediction. Early in sampling, DM retains indirect mistakes 15-20+ points more often than direct ones, whereas \ours shows only a 3.2-point difference on Sudoku and zero or even negative difference on mazes. On Sudoku, the gap for DM closes only around the 50\% sampling mark, when the violation becomes visually apparent. 
We take this as evidence that the diffusion model struggles most with mistakes that must be inferred rather than seen, whereas \ours handles both kinds similarly.

\begin{figure}[t]
    \centering
    \begin{minipage}[t]{0.485\linewidth}
        \centering
        {\footnotesize\hspace*{18pt}\textbf{Sudoku}}\\[2pt]
        \includegraphics[width=\linewidth]{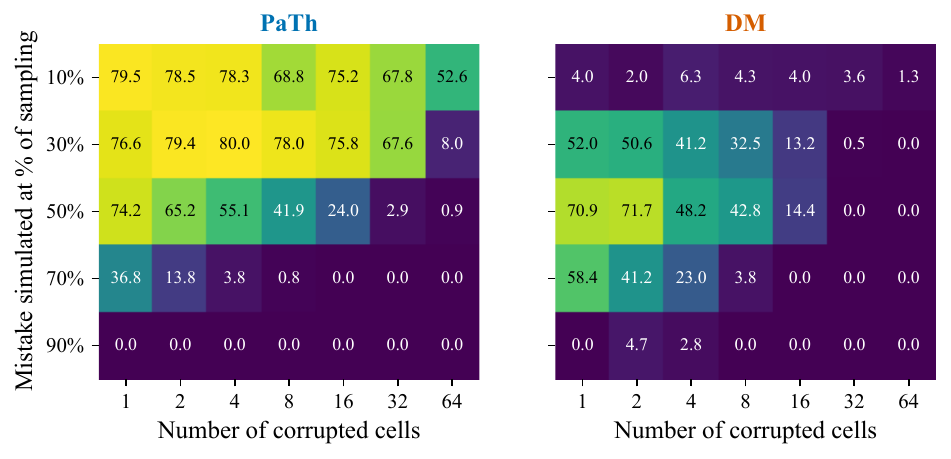}
    \end{minipage}
    \hfill
    \begin{minipage}[t]{0.485\linewidth}
        \centering
        {\footnotesize\hspace*{16pt}\textbf{CLEVR}}\\[2pt]
        \includegraphics[width=\linewidth]{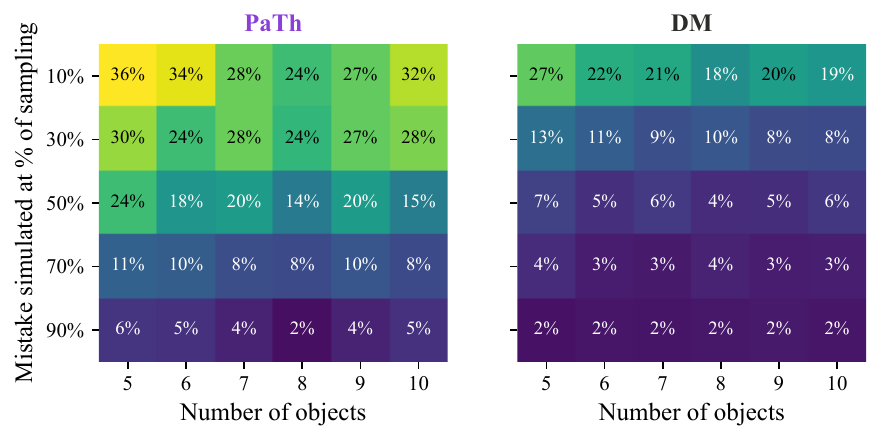}
    \end{minipage}
    \caption{Recovery rate after corrupting a valid solution and re-noising it
    to a given point of the trajectory, on \textsc{extreme} MNIST Sudoku
    (left) and constrained scenes (right). The diffusion model recovers only within a narrow band, at the start of
    sampling on CLEVR and at $50\%$ on Sudoku, and degrades sharply as mistakes
    accumulate. \ours recovers throughout the region where change remains
    possible, declining only as less trajectory is left.}
    \label{fig:fix_mistakes_recovery_rates}
    \vspace{-1em}
\end{figure}

\begin{wrapfigure}[14]{r}{0.38\linewidth}
\vspace{-0.5em}
    \centering
    \includegraphics[height=4cm]{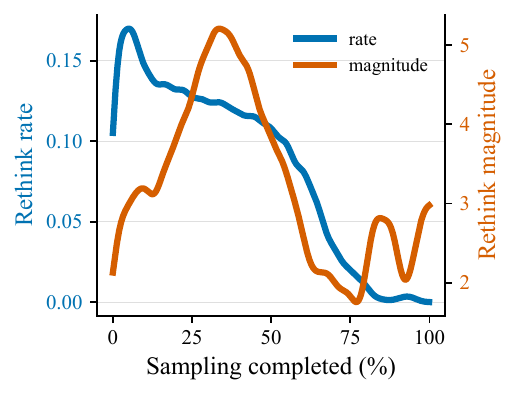}
    \vspace{-1.2em}
    \caption{
    Rethink rate (fraction of samples) and magnitude (violations) during sampling.}
    \label{fig:z_val_constraint_dip}
    \vspace{-1em}
\end{wrapfigure}
\paragraph{Probing the Thinker's latent state.} Finally, we propose to mechanistically validate what governs the decisions made by Thinker. 
To answer this question, we train a probe to decode the Thinker's current prediction from its internal state, allowing the state to be read as a candidate solution at every recursion (see App.~\ref{app:probe_training} for details). Over $384$ samples, we decode every state and count the constraints its prediction violates, giving a trajectory of violation counts within each denoising step. We say the model \emph{rethinks} at a step 
when this trajectory rises above its starting value, and report the rethink rate, the fraction of samples in which this occurs, together with the rethink magnitude, the mean size of the largest such rise.
The probe reveals a clear pattern of revision during the early stages of sampling (Fig.~\ref{fig:z_val_constraint_dip}). In up to a fifth of samples, then, the Thinker passes through candidates substantially worse than the one it began the step with, and does so precisely where most of the rule is resolved. This suggests that the latent trajectory is not merely refining an already-correct hypothesis: it sometimes explores less promising candidates before recovering and improving upon them. Fig.~\ref{fig:laten_reasoning} illustrates this behaviour on an individual puzzle, where successive recursions introduce additional violations before converging to a valid solution.

\begin{wrapfigure}{r}{0.44\linewidth}
\vspace{-3em}
    \centering
    \includegraphics[width=\linewidth]{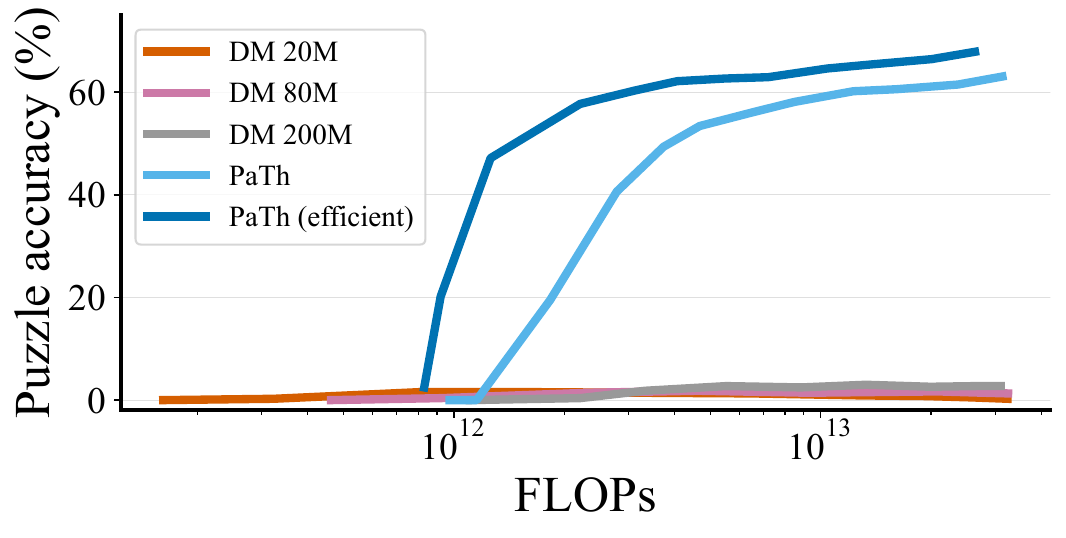}
    \vspace{-0.75\baselineskip}
    \caption{Puzzle accuracy on Sudoku \textsc{extreme} against inference FLOPs. DM stay below
    $5\%$ at every budget, whether the compute is spent on denoising steps or
    parameters, while \ours exceeds $60\%$. 
    }
    \label{fig:flops_scaling_results}
    \vspace{-0.75\baselineskip}
\end{wrapfigure}

\begin{figure}
\vspace{-1em}
    \centering
    \includegraphics[width=1.0\linewidth]{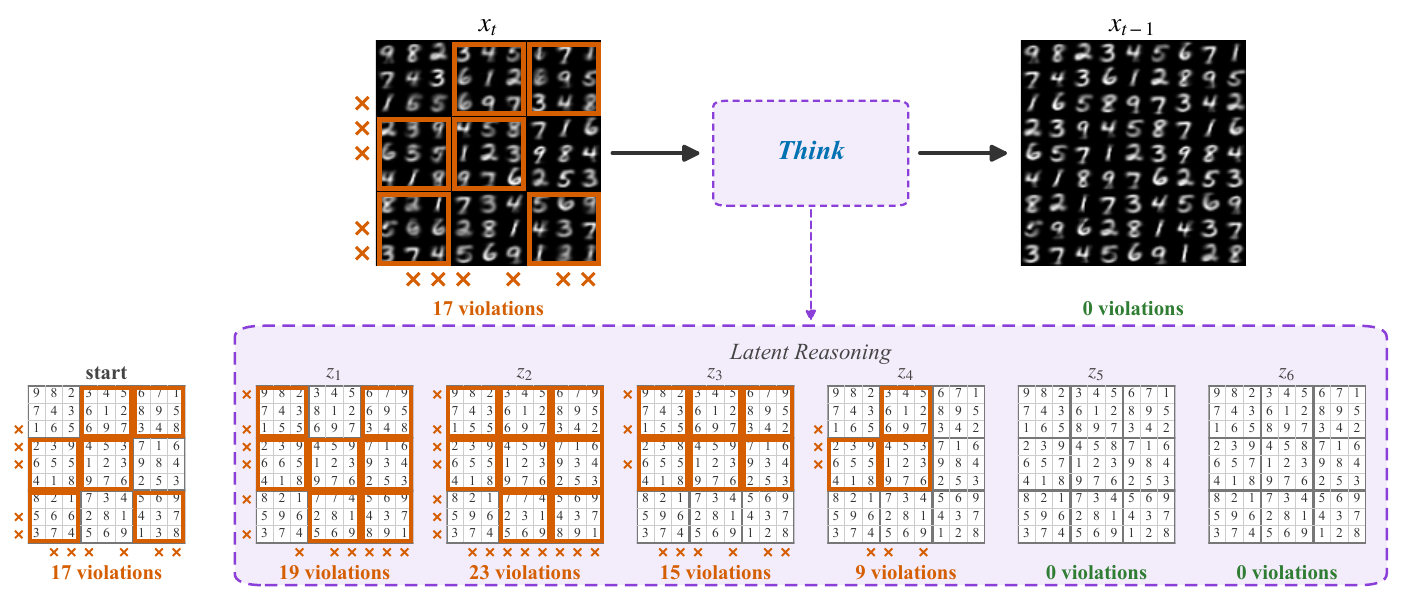}
    \caption{Latent reasoning within one denoising step, each recursion's state
$\mathbf{z}_i$ decoded by a linear probe with violated constraints in orange.
The Thinker introduces violations absent from $\mathbf{x}_t$ and then resolves
them, reaching a valid grid before anything is committed to
$\mathbf{x}_{t-1}$.}
 \label{fig:laten_reasoning}
\vspace{-1em}
\end{figure}

\vspace{-0.1em}
\subsection{Inference compute}
\vspace{-0.2em}
\label{sec:compute}
To evaluate if the reported improvement might simply reflect
a larger budget, we compare three ways of spending it: additional
recursions in the Thinker, additional denoising steps, and additional parameters
in the diffusion model. Fig.~\ref{fig:flops_scaling_results} plots puzzle
accuracy against inference FLOPs for each.
Diffusion models of $20$M, $80$M and $200$M parameters stay below $5\%$ accuracy over the entire range, whether the additional FLOPs go to more denoising steps or more parameters. \ours rises from near zero to over $60\%$ over the same range, starting at around $10^{12}$ FLOPs. Crucially, the leftmost point of the \ours curve is the trained model evaluated with a single recursion ($N_{\text{sup}}=H=L=1$), showing that the added module does not help by itself, and the gain comes from the recursion it performs.


Carrying the Thinker's
state across denoising steps rather than rebuilding it, concentrating the
recursion budget early, and omitting reasoning altogether once half the
trajectory has elapsed each helps, and each is consistent with the diagnosis
above: the state accumulates knowledge worth preserving, and most of the rule is
resolved in the first few steps. Together they give the efficient variant of
Fig.~\ref{fig:flops_scaling_results}, which reaches $47\%$ accuracy at $10^{12}$
FLOPs, where the standard configuration reaches approximately zero. We report
these ablations in App.~\ref{app:efficiency}.

\vspace{-0.3em}
\section{Conclusion}
\vspace{-0.2em}
Diffusion models that satisfy rules readily over symbols struggle when the same rules must hold in pixels, where every intermediate state is a partial image rather than a place to work out a solution. We show that a TRM-style recursive reasoner, run within each denoising step and coupled to a frozen diffusion model, is an effective way to add this reasoning capacity. Trained with a reconstruction loss alone, \ours improves over diffusion baselines across four families of rule-governed data, most on the hardest instances, at a fraction of the parameters. Our diagnostics suggest that the gain comes from the latent state and the recursion of the added module. 

%% file: appendix.tex
\section{Datasets}
\label{app:datasets}
\paragraph{MNIST Sudoku.} Each sample is a $9\times9$ grid of MNIST digits ($252\times252$ pixels), valid when each row, column, and $3\times3$ subgrid contains digits 1--9 exactly once. By default, we train at $144\times144$ resolution verifying full-resolution performance in App.~\ref{app:full_resolution_sudoku}. Given an incomplete grid, the model must fill the missing cells. We consider \textsc{hard}, where 0--26 cells are randomly revealed and multiple solutions are typically possible~\citep{wewer2025srm}, and \textsc{extreme}, using puzzles from TRM with roughly 18 clues and a unique solution requiring non-trivial reasoning. We report puzzle accuracy.

\paragraph{AMAZE benchmark.} We use both AMAZE tasks~\citep{zhou2026probing}. For mazes, the model must draw the path from start to goal while preserving the maze; for Queens, it must place one queen per row, column, and coloured region with no adjacent queens. Mazes range from $5\times5$ to $16\times16$ across four geometries, and Queens boards from $5\times5$ to $10\times10$. We report \textsc{coverage}, \textsc{violation}, $\textsc{pass}=\max(0,\textsc{coverage}-\textsc{violation})$, and $\textsc{exact}=\mathbf{1}_{\{\textsc{coverage} = 100 \land \textsc{violation} = 0\}}$ using @$1$ and @$5$ for \textsc{pass} and \textsc{exact}.

\paragraph{Constrained scene composition.} Based on CLEVR~\citep{johnson2017clevr}, scenes contain 3--10 objects at $256\times256$ resolution. The model receives admissible locations, object identities, and sparse pairwise relations (\emph{left of}, \emph{right of}, \emph{behind}, \emph{in front}) and must compose them transitively. We report spatial accuracy, precision, recall, and attribute binding; see App.~\ref{app:scene_eval} for definitions. Additional datasets are evaluated in App.~\ref{app:geosolvers_results}.

\subsection{Datasets - visual examples}
\label{app:ds_examples_imgs}
\begin{figure}[H]
    \centering
    \includegraphics[width=1.0\linewidth]{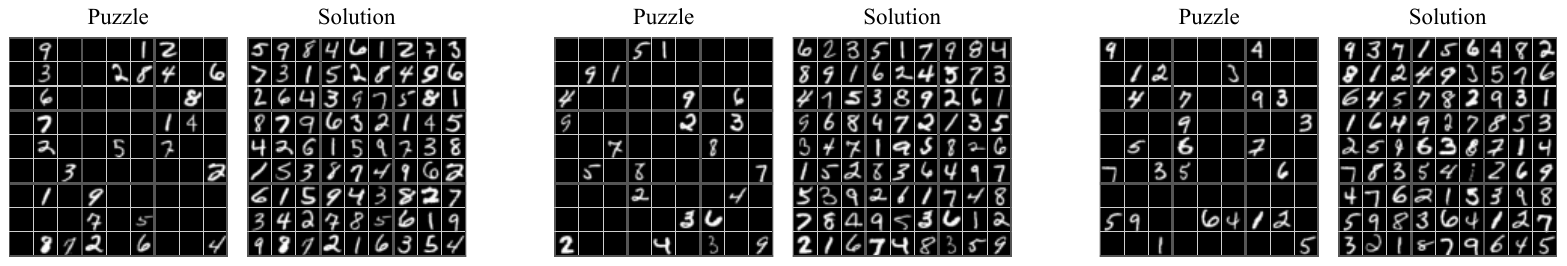}
    \caption{MNIST Sudoku puzzle examples.}
    \label{fig:sudoku_examples}
\end{figure}

\begin{figure}[H]
    \centering
    \includegraphics[width=1.0\linewidth]{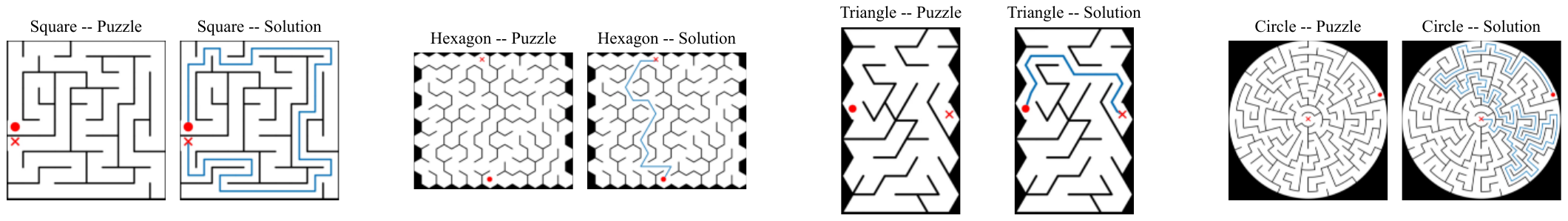}
    \caption{Mazes puzzle examples for all shapes.}
    \label{fig:mazes_examples}
\end{figure}

\begin{figure}[H]
    \centering
    \includegraphics[width=1.0\linewidth]{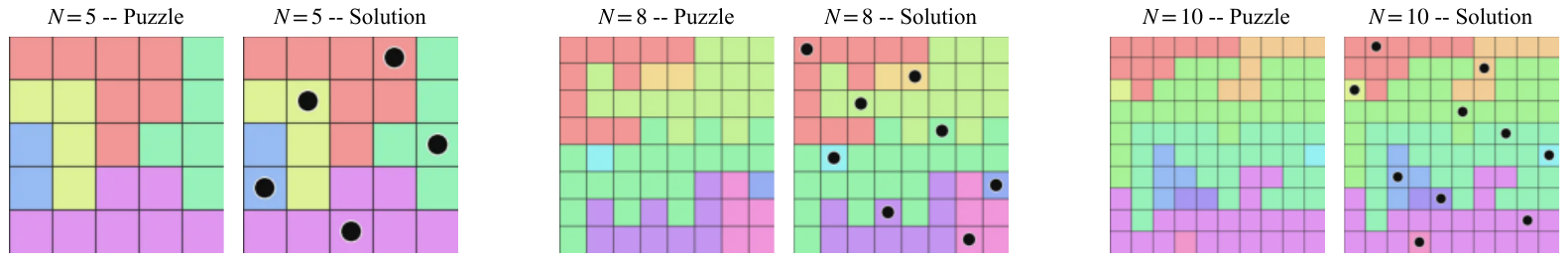}
    \caption{N-Queens puzzle examples for different sizes.}
    \label{fig:nqueens_examples}
\end{figure}

\begin{figure}[H]
    \centering
    \includegraphics[width=1.0\linewidth]{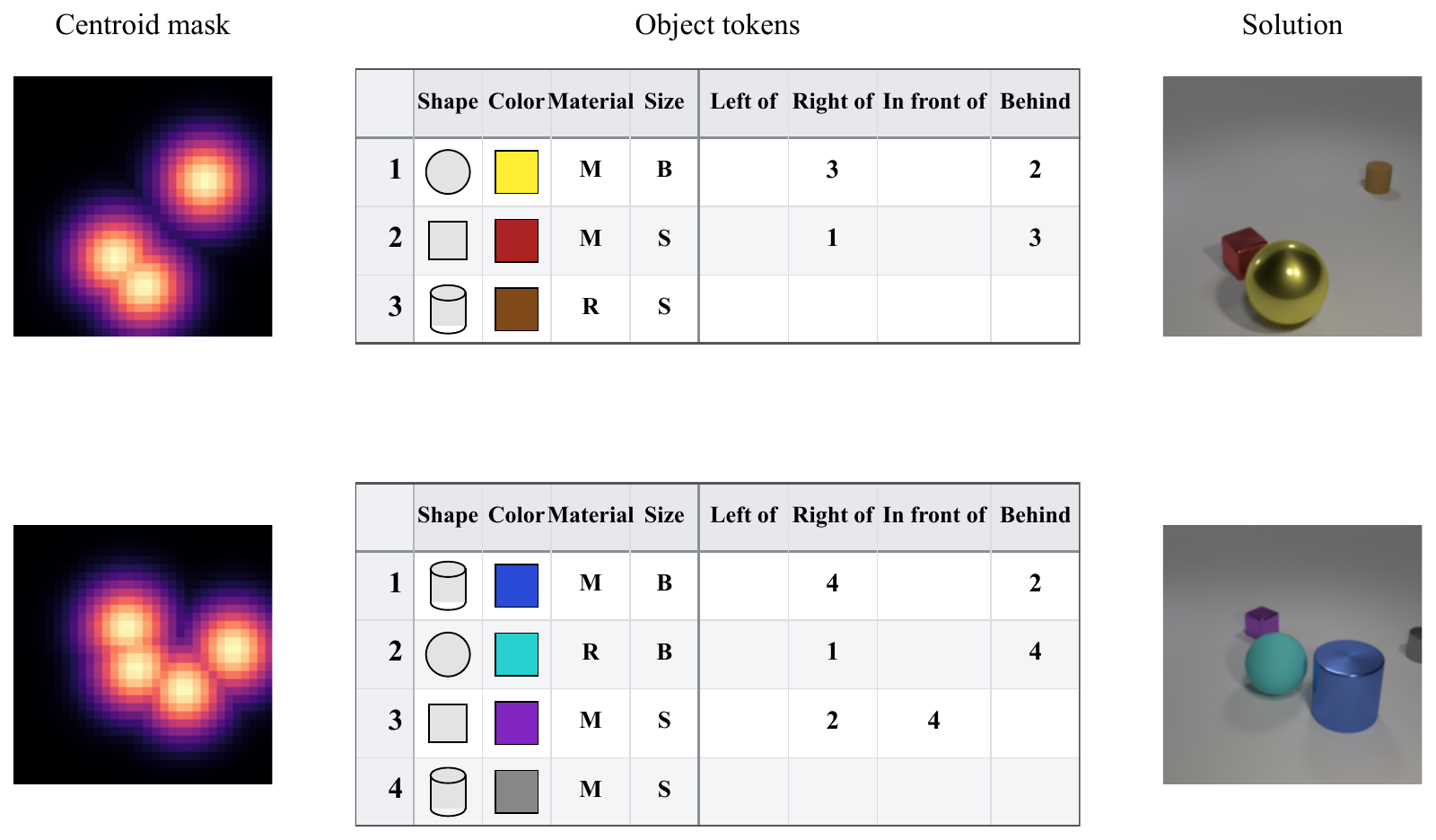}
    \caption{Constrained scene generation puzzle examples. Object tokens visualized for readability.}
    \label{fig:clevr_examples}
\end{figure}

We show conditioning and ground-truth solution pairs for each dataset: MNIST
Sudoku
(Fig.~\ref{fig:sudoku_examples}), mazes across the four grid geometries
(Fig.~\ref{fig:mazes_examples}), Queens boards
(Fig.~\ref{fig:nqueens_examples}), and constrained scene composition on CLEVR
(Fig.~\ref{fig:clevr_examples}).



\section{Scene composition evaluation method}
\label{app:scene_eval}

We evaluate generated images against their conditioning scene graph (a list
of ground-truth objects and pairwise spatial relations) with an
off-the-shelf open-vocabulary detect-and-attribute pipeline, applied
identically to every model.

\paragraph{Calibration.}
CLEVR relations are defined in 3-D world coordinates, while detections live
in image pixels. Using $150$ scenes from the official CLEVR validation set
(disjoint from any evaluated samples), we fit a homography from pixel-space
box centers to ground-plane $(x,y)$ coordinates, derive a small/large size
threshold from the two known object widths, and estimate unit ``left'' and
``front'' direction vectors by averaging the 3-D displacement between
objects in all annotated \texttt{left}/\texttt{front} relations. A relation
between two detected objects is later verified by checking the sign of the
dot product between their projected displacement and the corresponding
direction vector.

\paragraph{Detection.}
Each image is queried with Grounding~DINO~\cite{liu2024grounding} using a
text prompt listing every ground-truth object as a
\texttt{"\{size\} \{color\} \{material\} \{shape\}"} phrase, plus the three
bare shape names as a fallback. Detections are filtered with NMS
(IoU~$0.65$) and a simple center-distance de-duplication step to remove
near-duplicate boxes.

\paragraph{Attribute classification.}
For each detected box, we classify color, material, and shape jointly by
embedding two crops (a padded full-box crop and a center crop) with
SigLIP~\cite{zhai2023sigmoid} and taking the arg-max similarity against
text embeddings for all (color, material, shape) combinations, averaged
across the two crops. Size is assigned from the homography-projected box
width against the calibrated threshold.

\paragraph{Matching and hallucinations.}
Ground-truth and detected objects are matched via the Hungarian algorithm
under a cost equal to the number of mismatched attributes (color, shape,
material, size). A pair is accepted as a match only if at most $2$ of $4$
attributes disagree; detections left unmatched are counted as
hallucinations. Duplicate ground-truth objects sharing identical attributes
are disambiguated by trying all permutations of their matched positions and
keeping the one that satisfies the most ground-truth relations, so
attribute ties do not spuriously penalize relational accuracy.

\paragraph{Metrics.}
Let $t_{\text{req}}$/$t_{\text{pred}}$ be the number of requested/detected
objects, $v$ the number of accepted matches, and $t_{\text{rel}}$/$c_{\text{rel}}$
the number of checkable/satisfied ground-truth relations (both endpoints
matched). We report:
\begin{itemize}
    \item \textbf{Precision} $= v/t_{\text{pred}}$ and \textbf{Recall} $= v/t_{\text{req}}$ — conditioning adherence;
    \item \textbf{Color/Shape/Material/Size Acc.} — per-attribute agreement rate among matched objects;
    \item \textbf{Attribute binding} — fraction of matched objects correct on all four attributes;
    \item \textbf{Relational Acc.} $= c_{\text{rel}}/t_{\text{rel}}$ — fraction of spatial relations satisfied.
\end{itemize}
Counts are accumulated over the whole evaluation set before ratios are
taken. Results on the training-data samples are reported in
Table~\ref{tab:app_clevr_eval_on_train}.

\begin{table}[H]
    \centering
    \begin{tabular}{c c c c c c c c}
        \toprule
        \multicolumn{2}{c}{General adherence} & \multicolumn{5}{c}{Attribute matching} & \multirow{2}{*}{\shortstack{Spatial \\ relations \\ accuracy}} \\
        \cmidrule(lr){1-2} \cmidrule(lr){3-7}
        recall & precision & color & shape & material & size & all & \\
        \midrule
        99.9 & 99.8 & 95.2 & 96.9 & 96.4 & 100 & 89.5 & 97.2 \\
        \bottomrule
    \end{tabular}
    \caption{Performance of the evaluation method on CLEVR ground truth training data.}
    \label{tab:app_clevr_eval_on_train}
\end{table}

\section{Implementation details}
\label{app:implementation_details}
Tab.~\ref{tab:hyperparameters} lists all hyperparameters. The Painter is a UNet
denoiser predicting $\mathbf{x}_0$ over $100$ training timesteps, sampled in
$20$ steps with classifier-free guidance at scale $2.0$, and maintained as an
EMA with rate $0.999$. On CLEVR it operates in the latent space of a pretrained
SDXL VAE with cross-attention blocks; elsewhere it works directly in pixel
space. The Thinker follows \citet{jolicoeurmartineau2025trm} with two
self-attention layers, hidden size $512$, eight heads and rotary position
encodings, run with $H = 3$, $L = 6$ and $N_{\text{sup}} = 16$. The Painter is
trained first and then frozen, after which the Thinker and the condition encoder
are trained under the reconstruction loss alone. The parameter counts for all models and for each dataset are provided in Tab.~\ref{tab:param_counts}.
\begin{table}[H]
    \centering
    \vspace{-\intextsep}
    \resizebox{\linewidth}{!}{
        \begin{tabular}{l c c c c}
        \toprule
        Dataset & Painter & Thinker & \ours & DM \\
        \midrule
        MNIST Sudoku  & 1.5M & 8.3M & 10.3M & 81.6M\\
        Mazes  & 1.5M & 8.3M & 10.3M & 62.7M \\
        Queens  & 1.5M & 8.3M & 10.3M & 62.7M \\
        Constrained scenes  & 101.9M & 19.7M & 121.6M & 236.0M \\
        \bottomrule
        \end{tabular}
    }
    \caption{Parameter counts by dataset. \ours is the sum of its Painter,
Thinker, conditioning encoders, and ControlNet adapters; DM is the diffusion baseline \vspace{-1em}}
    \label{tab:param_counts}
\end{table}

\begin{table}[t]
    \centering
    \setlength{\tabcolsep}{3pt}
    \small 
    \begin{tabular}{l l l l l}
        \toprule
        \textbf{Parameter} & \textbf{MNIST Sudoku} & \textbf{CLEVR} & \textbf{Mazes} & \textbf{N-Queens} \\
        \midrule
        
        \multicolumn{5}{l}{\textbf{Painter}} \\
        \midrule
        VAE & - & stabilityai/sdxl-vae & - & - \\
        Block types & Block2D $\times 3$ & CrossAttBlock $\times 3$ & Block2D $\times 3$ & Block2D $\times 3$ \\
        Block out channels & 32, 64, 64 & 128, 256, 512 & 32, 64, 64 & 32, 64, 64\\
        Cross att dim & - & 512 & - & - \\
        Layers per block & 2 & 2 & 2 & 2 \\
        Painter training cfg & 0 & 0.9 & 0 & 0 \\
        \addlinespace
        
        \multicolumn{5}{l}{\textbf{Painter Optimizer}} \\
        \midrule
        Optim type & AdamW & AdamW & AdamW & AdamW \\
        Learning rate (lr) & 3e-5 & 3e-5 & 1e-4 & 1e-4 \\
        lr min ratio & 0.1 & 0.1 & 1 & 1 \\
        Warmup steps & 1000 & 1000 & 2000 & 2000 \\
        Weight decay & 0 & 0 & 0.1 & 0.1 \\
        Training steps & 100000 & 80000 & 40000 & 40000 \\
        \addlinespace
        
        \multicolumn{5}{l}{\textbf{Condition Encoder}} \\
        \midrule
        Type & conv & conv+mlp & conv & conv \\
        Enc channels & 128 & 55 & 128 & 128 \\
        Hidden channels & 128, 256, 256 & 512 & 128, 256, 256 & 128, 256, 256 \\
        Output dim & 512 & 512 & 512 & 512 \\
        \addlinespace
        
        \multicolumn{5}{l}{\textbf{Diffusion}} \\
        \midrule
        Num train timesteps & 100 & 100 & 100 & 100 \\
        Prediction type & x0 & x0 & x0 & x0 \\
        Num sampling timesteps & 20 & 20 & 20 & 20 \\
        Sampling cfg & 2.0 & 2.0 & 2.0 & 2.0 \\
        \addlinespace
        
        \multicolumn{5}{l}{\textbf{EMA}} \\
        \midrule
        Rate & 0.999 & 0.999 & 0.999 & 0.999 \\
        \addlinespace
        
        \multicolumn{5}{l}{\textbf{Thinker}} \\
        \midrule
        Expansion factor & 4 & 4 & 4 & 4 \\
        Hidden size & 512 & 512 & 512 & 512 \\
        Num layers & 2 & 2 & 2 & 2 \\
        Att num heads & 8 & 8 & 8 & 8 \\
        Pos encoding & rope & rope & rope & rope \\
        Sequence length & 81 & 266 & 144 & 144 \\
        Out shape & 11 & 10 & 11 & 11 \\
        H cycles & 3 & 3 & 3 & 3 \\
        L cycles & 6 & 6 & 6 & 6 \\
        N sup & 16 & 16 & 16 & 16 \\
        \addlinespace
        
        \multicolumn{5}{l}{\textbf{Thinker Optimizer}} \\
        \midrule
        Optim type & AdamAtan2 & AdamAtan2 & AdamAtan2 & AdamAtan2 \\
        Learning rate (lr) & 1e-4 & 1e-4 & 1e-4 & 1e-4 \\
        lr min ratio & 1 & 1 & 1 & 1 \\
        Warmup steps & 2000 & 2000 & 2000 & 2000 \\
        Weight decay & 0.3 & 0.3 & 0.3 & 0.3 \\
        Betas & 0.9, 0.95 & 0.9, 0.95 & 0.9, 0.95 & 0.9, 0.95 \\
        Training steps & 160000 & 90000 & 40000 & 40000 \\
        \bottomrule
    \end{tabular}
        \caption{Hyperparameters across Components and Tasks}
    \label{tab:hyperparameters}
\end{table}

\section{Additional results - geosolvers datasets}
\label{app:geosolvers_results}
We additionally evaluate on the three tasks of \citet{goren2026geometric}, in
which a diffusion model trained in pixel space approximates solutions to the
Simple Polygon, Steiner Tree and Inscribed Square problems. Following their
protocol, we train only on the smallest instances and evaluate across all sizes.

\ours improves substantially on Simple Polygon
(Tab.~\ref{tab:results_polygons}), raising validity from $0.620$ to $0.869$ and
the optimality rate from $0.062$ to $0.264$ on the largest instances. On Steiner
Tree (Tab.~\ref{tab:results_steiner}) and Inscribed Square
(Tab.~\ref{tab:results_square}) the two models are comparable, with \ours
slightly ahead on solution ratio and slightly behind on validity. We attribute
this to the nature of the tasks rather than to headroom: Simple Polygon is a
discrete constraint satisfaction problem, where a configuration is valid or not,
whereas the other two are continuous optimisation problems whose solutions are
approximate by construction. The Thinker searches over candidates that can be
rejected, which is what the former requires and the latter does not.
\begin{table}[H]
    \centering
    \begin{tabular}{l c c c c c c}
        \toprule
        \multirow{2}{*}{Model} & \multicolumn{3}{c}{7-12} & \multicolumn{3}{c}{13-15} \\
        \cmidrule(lr){2-4} \cmidrule(lr){5-7}
        & valid & ratio & opt rate & valid & ratio & opt rate \\
        \midrule
        \ours & \textbf{0.985} & \textbf{0.998} & \textbf{0.783} & \textbf{0.869} & \textbf{1.000} & \textbf{0.264} \\
        visual geosolvers & 0.953 & 0.988 & 0.574 & 0.620 & 0.962 & 0.062 \\
        \bottomrule
    \end{tabular}
    \caption{Simple Polygon, by number of vertices. \emph{valid} is the fraction of
outputs forming a simple polygon, \emph{ratio} their quality relative to the
reference, and \emph{opt rate} the fraction attaining the optimum.}
    \label{tab:results_polygons}
\end{table}
\begin{table}[H]
    \centering
    \begin{tabular}{l c c c c c c c c}
        \toprule
        \multirow{2}{*}{Model} & \multicolumn{2}{c}{10-20} & \multicolumn{2}{c}{21-30} & \multicolumn{2}{c}{31-40} & \multicolumn{2}{c}{41-50} \\
        \cmidrule(lr){2-3} \cmidrule(lr){4-5} \cmidrule(lr){6-7} \cmidrule(lr){8-9}
        & valid & ratio & valid & ratio & valid & ratio & valid & ratio \\
        \midrule
        \ours & \textbf{0.997} & \textbf{1.0008} & 0.983 & \textbf{1.0018} & 0.796 & \textbf{1.0040} & \textbf{0.340} & \textbf{1.0081} \\
        visual geosolvers & 0.996 & \textbf{1.0008} & \textbf{0.986} & \textbf{1.0018} & \textbf{0.834} & 1.0044 & 0.334 & 1.0092 \\
        \bottomrule
    \end{tabular}
    \caption{Steiner Tree, by number of terminals. Lower \emph{ratio} is better.
The two models are comparable throughout; neither solves the largest instances.}
    \label{tab:results_steiner}
\end{table}
\begin{table}[H]
    \centering
    \begin{tabular}{l c c}
        \toprule
        Model & Squareness & Alignment \\
        \midrule
        \ours & 0.874 & \textbf{-0.84} \\
        visual geosolvers & \textbf{0.891} & -0.90 \\
        \bottomrule
    \end{tabular}
    \caption{Inscribed Square. Neither model has a consistent advantage.}
    \label{tab:results_square}
\end{table}

\section{Stability of the method}
\label{app:stability}
\ours is stable across seeds. Three runs on \textsc{extreme} MNIST Sudoku give
puzzle accuracies of $71.2$, $70.8$ and $69.8$, a mean of $70.6 \pm 0.7$.

It is also insensitive to the Thinker's token grid
(Tab.~\ref{tab:grid_sizes}). Grids of $10 \times 10$ and $12 \times 12$ neither
align with the $9 \times 9$ Sudoku cells nor divide into them, yet reach $66.8$
and $69.3$ against $71.2$ for the aligned $9 \times 9$. The method therefore
requires no correspondence between the Thinker's resolution and the structure of
the task.

\begin{table}[H]
    \centering
    \begin{tabular}{l c c c c}
        \toprule
        TRM num tokens & 8x8 & 9x9 & 10x10 & 12x12 \\
        \midrule
        puzzle acc & 59.8 & \textbf{71.2} & 66.8 & 69.3 \\
        \bottomrule
    \end{tabular}
    \caption{Puzzle accuracy on \textsc{extreme} MNIST Sudoku for different Thinker
token grids. Only $9 \times 9$ aligns with the Sudoku cells.}
    \label{tab:grid_sizes}
\end{table}

\section{Adoption to DiT}
\label{app:adoption_to_dit}
\ours does not depend on the Painter being a UNet. A DiT has no skip
connections to inject into, so in place of the ControlNet adapters we project
the Thinker's output linearly and add it token-wise to the hidden states after
each transformer block. The remainder of the setup is unchanged.

Tab.~\ref{tab:dit_results} reports the result. The DiT variant reaches $0.542$
puzzle accuracy against $0.029$ for the corresponding diffusion baseline, so the
mechanism transfers, though it falls short of the $0.712$ obtained with a UNet
Painter. Since the Thinker is identical in both cases, we attribute the gap to
the coupling, which we did not tune further.

\begin{table}[H]
    \centering
    \begin{tabular}{l c}
        \toprule
        Painter architecture & Puzzle accuracy  \\
        \midrule
        UNet & 0.712  \\
        DiT & 0.542 \\
        DM & 0.029 \\ 
        \bottomrule
    \end{tabular}
    \caption{Puzzle accuracy on \textsc{extreme} MNIST Sudoku with different Painter
architectures. DM is a diffusion model without a Thinker.}
\label{tab:dit_results}
\end{table}

\section{Halting head impact}
\label{app:halting_head_training}

The halting head reduces the cost of Thinker training by avoiding recursive
supervision steps that are predicted to provide no further improvement. On
MNIST Sudoku, training the Thinker without halting takes $72$h, whereas the
halting variant takes only $26$h, reducing the total two-stage training cost
from $85$h to $39$h when combined with the $13$h Painter training stage.

The reduction does not come from a higher throughput. The iteration rate is
essentially unchanged, from $0.85$ iterations/s without halting to
$0.81$ iterations/s with halting. Instead, the halting head reduces the number
of recursive supervision steps performed during training by terminating
recursion once an additional step is predicted to yield no further
improvement. The resulting savings therefore come from avoiding unnecessary
computation rather than making individual iterations cheaper.

\begin{table}[H]
\centering
\begin{tabular}{l c c}
\toprule
Configuration & iter/s & Training time \\
\midrule
Separate, Painter & 5.8 & 13h \\
Separate, Thinker & 0.85 & 72h \\
Separate, total & -- & 85h \\
\midrule
Separate, Thinker with halting & 0.81 & 26h \\
Separate, total with halting & -- & 39h \\
\bottomrule
\end{tabular}
\caption{Training cost on MNIST Sudoku. Total training times sum the
Painter and Thinker stages; iteration rates are not defined for the
combined two-stage configurations.}
\label{tab:efficiency}
\end{table}

\section{Joint vs separate training}
\label{app:joint_vs_separate}

We compare joint training of the Painter and Thinker with our two-stage
procedure in which the Painter is trained first and then frozen while the
Thinker is trained. As shown in Tab.~\ref{tab:joint_vs_separable}, separate
training requires $85$h compared with $113$h for joint training, while also
achieving slightly higher puzzle accuracy on MNIST Sudoku: $71.2\%$ versus
$69.9\%$.

\begin{figure}[H]
\centering
\includegraphics[width=\linewidth]{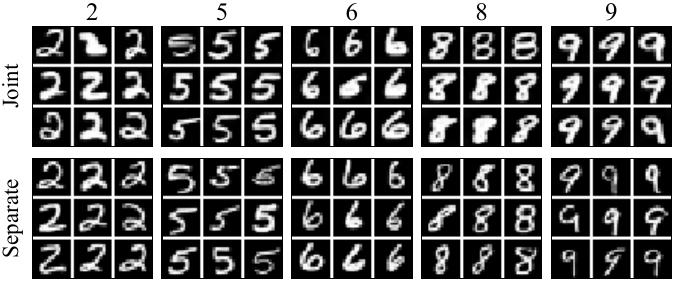}
\caption{Examples generated by jointly and separately trained
\ours models. Separately trained models exhibit greater
diversity in the generated MNIST digits.}
\label{fig:joint_vs_separate}
\end{figure}

The qualitative difference is also visible in Fig.~\ref{fig:joint_vs_separate}.
The separately trained model produces a more diverse set of MNIST digits,
whereas joint training produces samples with less variation. This difference
suggests that keeping the Painter fixed allows the Thinker to adapt its
conditioning signal to a stable image-generation model without jointly
changing the underlying digit distribution.

To quantify this, we split the generated $9\times9$ grids of two test puzzles
into their cells and label each cell by the digit it depicts. For each digit class we compute the
mean pairwise $\ell_2$ distance between samples, with each sample flattened and
scaled to unit norm. We then average this value over the nine classes. The
separate model is more diverse in all nine classes ($0.855$ vs.\ $0.721$,
Tab.~\ref{tab:joint_vs_separable}).

\begin{table}[H]
\centering
\begin{tabular}{l c c c}
\toprule
Training configuration & Training time & Puzzle accuracy & Digit diversity $\uparrow$ \\
\midrule
Joint & 113h & 69.9\% & 0.721 \\
Separate & 85h & 71.2\% & 0.855 \\
\bottomrule
\end{tabular}
\caption{Comparison of joint and separate Painter--Thinker training on
MNIST Sudoku. Separate training is less costly, slightly more accurate and
produces more diverse digits in this setting. Digit diversity is the
within-class mean pairwise $\ell_2$ distance between unit-normalized samples,
averaged over digit classes.}
\label{tab:joint_vs_separable}
\end{table}

\section{ControlNet alternative}
\label{app:concat_vs_controlnet}
We also evaluate a simpler alternative to the ControlNet coupling in which the
Thinker's output is concatenated channel-wise with the diffusion state
$\mathbf{x}_t$. We first train the diffusion model on full solved MNIST Sudoku
grids, with the puzzle conditioning concatenated directly to $\mathbf{x}_t$.
During subsequent Thinker training, the Thinker's output is used as the
conditioning signal and is likewise concatenated with $\mathbf{x}_t$ before
being passed to the diffusion model.

On MNIST Sudoku \textsc{extreme}, this concatenation-based variant reaches
$37.5\%$ puzzle accuracy, compared with $71.2\%$ for the ControlNet-based
model. The concatenation approach therefore retains a substantial benefit
over diffusion baselines, but performs considerably worse than ControlNet.
These results indicate that the interface between the Thinker and Painter is
important: injecting the Thinker's representation through the Painter's
ControlNet adapters is substantially more effective than treating it as an
additional input channel.

\section{Recovering mistakes: full metrics}
\label{app:full_recover_mistakes}

The main text reports the recovered-and-valid rate, which requires the corrupted
cells to be restored and the rest of the grid to remain correct. Here we report
the recovery rate without the second requirement, counting only whether the
corrupted cells themselves return to the ground truth, together with the kept
rate, the fraction of corruptions left unchanged. The two do not sum to one,
since a corrupted cell may be altered to a third value. Results are separated by
whether a corruption is locally visible, duplicating a digit in its own row,
column or block, or globally visible, creating no such duplication while still
departing from the unique solution.

Comparing the two recovery criteria is informative. At $30\%$ of sampling with
$32$ locally visible corruptions, the diffusion model restores $58.8\%$ of the
corrupted cells but produces a valid grid in only $7.0\%$ of instances: it
repairs what it was given and disturbs the rest. \ours restores $97.7\%$ and
keeps $74.7\%$ of grids valid.

\begin{table}[H]
\centering
\resizebox{\textwidth}{!}{
\begin{tabular}{ccccccccccccccc}
\toprule
 & \multicolumn{7}{c}{\ours} & \multicolumn{7}{c}{DM} \\
\cmidrule(lr){2-8} \cmidrule(lr){9-15}
\% sampling & $n{=}1$ & $n{=}2$ & $n{=}4$ & $n{=}8$ & $n{=}16$ & $n{=}32$ & $n{=}64$ & $n{=}1$ & $n{=}2$ & $n{=}4$ & $n{=}8$ & $n{=}16$ & $n{=}32$ & $n{=}64$ \\
\midrule
10 & 94.5 & 96.5 & 96.1 & 94.3 & 94.5 & 94.2 & 86.8 & 49.2 & 50.0 & 49.6 & 52.2 & 50.5 & 49.0 & 44.7 \\
30 & 96.9 & 96.1 & 96.5 & 98.1 & 96.0 & 97.7 & 64.8 & 85.9 & 88.3 & 86.9 & 85.0 & 77.6 & 58.8 & 42.4 \\
50 & 89.1 & 95.3 & 91.2 & 90.1 & 90.4 & 90.6 & 45.9 & 97.7 & 96.9 & 96.5 & 95.8 & 92.0 & 74.2 & 36.8 \\
70 & 42.2 & 46.5 & 43.9 & 40.9 & 39.2 & 35.4 & 15.9 & 77.3 & 72.7 & 77.9 & 70.6 & 57.7 & 32.3 & 12.1 \\
90 & 0.0 & 0.4 & 0.4 & 0.3 & 0.3 & 0.3 & 0.3 & 10.9 & 7.0 & 8.2 & 6.9 & 6.3 & 3.6 & 1.9 \\
\bottomrule
\end{tabular}}
\caption{Recovery rate (\%) for locally visible corruptions on \textsc{extreme}
MNIST Sudoku, by percentage of sampling completed when the mistakes are
injected and number $n$ of corrupted cells. This does not require the rest of
the grid to remain correct.}
\label{tab:sudoku_recovered_local}
\end{table}

\begin{table}[H]
\centering
\resizebox{\textwidth}{!}{
\begin{tabular}{ccccccccccccccc}
\toprule
 & \multicolumn{7}{c}{\ours} & \multicolumn{7}{c}{DM} \\
\cmidrule(lr){2-8} \cmidrule(lr){9-15}
\% sampling & $n{=}1$ & $n{=}2$ & $n{=}4$ & $n{=}8$ & $n{=}16$ & $n{=}32$ & $n{=}64$ & $n{=}1$ & $n{=}2$ & $n{=}4$ & $n{=}8$ & $n{=}16$ & $n{=}32$ & $n{=}64$ \\
\midrule
10 & 94.5 & 95.2 & 93.3 & 93.7 & 92.4 & 91.5 & 76.7 & 48.4 & 51.2 & 55.9 & 50.6 & 49.1 & 45.9 & 40.9 \\
30 & 97.7 & 96.8 & 96.0 & 95.1 & 95.0 & 90.8 & 40.7 & 89.0 & 90.5 & 82.2 & 80.9 & 67.2 & 42.8 & 30.6 \\
50 & 92.2 & 87.1 & 90.0 & 90.5 & 90.7 & 69.4 & 26.5 & 95.3 & 95.3 & 93.7 & 90.3 & 78.9 & 39.6 & 19.9 \\
70 & 43.2 & 41.3 & 37.2 & 42.5 & 36.0 & 20.2 & 8.8 & 76.8 & 74.1 & 70.1 & 61.4 & 43.0 & 17.8 & 8.0 \\
90 & 0.0 & 0.0 & 0.2 & 0.3 & 0.6 & 0.2 & 0.3 & 7.1 & 12.7 & 5.7 & 7.1 & 4.7 & 2.2 & 1.1 \\
\bottomrule
\end{tabular}}
\caption{Recovery rate (\%) for globally visible corruptions.}
\label{tab:sudoku_recovered_global}
\end{table}

\begin{table}[H]
\centering
\resizebox{\textwidth}{!}{
\begin{tabular}{ccccccccccccccc}
\toprule
 & \multicolumn{7}{c}{\ours} & \multicolumn{7}{c}{DM} \\
\cmidrule(lr){2-8} \cmidrule(lr){9-15}
\% sampling & $n{=}1$ & $n{=}2$ & $n{=}4$ & $n{=}8$ & $n{=}16$ & $n{=}32$ & $n{=}64$ & $n{=}1$ & $n{=}2$ & $n{=}4$ & $n{=}8$ & $n{=}16$ & $n{=}32$ & $n{=}64$ \\
\midrule
10 & 0.8 & 0.0 & 0.0 & 0.2 & 0.1 & 0.1 & 0.2 & 0.0 & 0.0 & 0.2 & 0.2 & 0.3 & 0.2 & 0.3 \\
30 & 0.0 & 0.4 & 0.2 & 0.1 & 0.0 & 0.2 & 0.3 & 0.0 & 0.0 & 1.4 & 0.0 & 0.4 & 0.5 & 0.5 \\
50 & 4.7 & 1.6 & 3.3 & 4.2 & 4.3 & 4.0 & 6.6 & 1.6 & 0.0 & 1.2 & 0.9 & 1.0 & 1.5 & 2.7 \\
70 & 37.5 & 40.6 & 42.8 & 42.7 & 45.9 & 48.8 & 52.2 & 14.1 & 22.7 & 15.0 & 20.9 & 28.4 & 38.1 & 45.4 \\
90 & 94.5 & 98.8 & 97.7 & 97.5 & 97.4 & 97.6 & 97.5 & 86.7 & 88.3 & 87.1 & 89.4 & 88.4 & 90.3 & 90.2 \\
\bottomrule
\end{tabular}}
\caption{Kept rate (\%) for locally visible corruptions.}
\label{tab:sudoku_kept_local}
\end{table}

\begin{table}[H]
\centering
\resizebox{\textwidth}{!}{
\begin{tabular}{ccccccccccccccc}
\toprule
 & \multicolumn{7}{c}{\ours} & \multicolumn{7}{c}{DM} \\
\cmidrule(lr){2-8} \cmidrule(lr){9-15}
\% sampling & $n{=}1$ & $n{=}2$ & $n{=}4$ & $n{=}8$ & $n{=}16$ & $n{=}32$ & $n{=}64$ & $n{=}1$ & $n{=}2$ & $n{=}4$ & $n{=}8$ & $n{=}16$ & $n{=}32$ & $n{=}64$ \\
\midrule
10 & 3.9 & 1.2 & 2.8 & 2.6 & 3.0 & 3.1 & 8.7 & 15.1 & 16.5 & 19.6 & 18.5 & 18.4 & 20.6 & 21.7 \\
30 & 0.8 & 0.4 & 0.6 & 2.1 & 2.0 & 4.5 & 26.6 & 5.5 & 4.3 & 6.9 & 9.1 & 13.8 & 24.7 & 31.4 \\
50 & 3.9 & 6.6 & 4.7 & 3.7 & 4.6 & 16.5 & 40.7 & 2.4 & 2.8 & 3.9 & 6.0 & 12.6 & 33.9 & 43.9 \\
70 & 43.2 & 46.5 & 46.3 & 44.0 & 49.4 & 65.5 & 72.5 & 17.6 & 19.6 & 23.4 & 28.3 & 43.1 & 60.2 & 67.2 \\
90 & 97.7 & 98.0 & 98.6 & 97.8 & 97.4 & 97.7 & 97.8 & 88.2 & 82.1 & 90.2 & 87.1 & 91.0 & 93.6 & 94.4 \\
\bottomrule
\end{tabular}}
\caption{Kept rate (\%) for globally visible corruptions. The difference
between this table and Tab.~\ref{tab:sudoku_kept_local} is plotted in
Fig.~\ref{fig:keep_mistake_local_vs_global}.}
\label{tab:sudoku_kept_global}
\end{table}

For constrained scenes, a corruption exchanges the positions of $n$ disjoint
pairs of objects in a seven-object scene. We report the fraction of scenes in
which every swapped object is correctly placed, the per-object recovery rate,
and the per-object rate at which an object remains in its swapped position.

\begin{table}[H]
\centering
\begin{tabular}{ccccccc}
\toprule
 & \multicolumn{3}{c}{\ours} & \multicolumn{3}{c}{DM} \\
\cmidrule(lr){2-4} \cmidrule(lr){5-7}
\% sampling & $n{=}1$ & $n{=}2$ & $n{=}3$ & $n{=}1$ & $n{=}2$ & $n{=}3$ \\
\midrule
10 & 53.2 & 55.7 & 53.0 & 40.7 & 41.4 & 41.1 \\
30 & 51.2 & 52.1 & 51.2 & 25.5 & 26.4 & 26.3 \\
50 & 39.8 & 41.4 & 39.7 & 14.2 & 14.7 & 15.5 \\
70 & 21.0 & 24.6 & 23.4 & 9.2 & 9.8 & 9.1 \\
90 & 13.0 & 12.2 & 12.6 & 6.0 & 7.6 & 6.8 \\
\bottomrule
\end{tabular}
\caption{Per-object recovery rate (\%) on constrained scenes.}
\label{tab:clevr_swap_success}
\end{table}

\begin{table}[H]
\centering
\begin{tabular}{ccccccc}
\toprule
 & \multicolumn{3}{c}{\ours} & \multicolumn{3}{c}{DM} \\
\cmidrule(lr){2-4} \cmidrule(lr){5-7}
\% sampling & $n{=}1$ & $n{=}2$ & $n{=}3$ & $n{=}1$ & $n{=}2$ & $n{=}3$ \\
\midrule
10 & 43.8 & 39.4 & 41.7 & 53.2 & 55.5 & 56.0 \\
30 & 44.0 & 41.7 & 43.5 & 71.0 & 70.1 & 70.8 \\
50 & 54.2 & 52.2 & 53.6 & 82.5 & 81.6 & 80.7 \\
70 & 74.0 & 71.1 & 71.6 & 87.7 & 87.5 & 87.5 \\
90 & 80.0 & 82.5 & 82.3 & 92.0 & 88.9 & 89.4 \\
\bottomrule
\end{tabular}
\caption{Per-object kept rate (\%) on constrained scenes.}
\label{tab:clevr_swap_stuck}
\end{table}

\section{Thinker probe training details}
\label{app:probe_training}
To inspect what the model knows at an intermediate point in its reasoning,
without going through the Painter, which blurs early noisy states and would
confound the reading, we train a small linear probe $p_\theta$ mapping a single
per-cell $\mathbf{y}$ vector to a nine-way digit distribution.

\paragraph{Training.} We sample puzzles, recording $\mathbf{y}$ at every
recursion of every denoising step. The label for all of these, regardless of
when they were captured, is a standard MNIST classifier's cell-by-cell reading
of that trajectory's own final image. $p_\theta$ is trained with cross-entropy
against this label.

\paragraph{Reading the latent.} The probe is defined on $\mathbf{y}$, since
$\mathbf{z}$ does not itself decode to a solution
\citep[Fig.~6]{jolicoeurmartineau2025trm}. To read a latent state
$\mathbf{z}_i$, we therefore apply the Thinker's answer update
$\mathbf{y} \leftarrow \mathrm{net}(\mathbf{y}, \mathbf{z}_i)$ and probe the
result, which reports the prediction the latent would produce were the model to
commit at that point.

\paragraph{Validation.} Tab.~\ref{tab:internal_state_acc} reports probe accuracy
by denoising step $t$ and number of completed supervision steps, read out
continuously within each denoising step. From $t{=}5$ onwards accuracy exceeds
$95\%$ and is essentially unchanged by further reasoning, so the probe recovers
the committed digit well before the image is rendered. Accuracy is lower at both
extremes. At $t{=}0$ the state has not yet converged, though it improves with
reasoning depth, from $66.2\%$ to $89.1\%$. At $t{=}19$ the sample is nearly
clean and the Thinker has little left to do, so $\mathbf{y}$ is no longer driven
towards the solution; this is consistent with
Fig.~\ref{fig:reuse_steering}, where reasoning in the final steps can be omitted
without loss. Our analyses use the intermediate range.

\begin{table}[H]
    \centering
    \begin{tabular}{l c c c c c}
        \toprule
        t \textbackslash{} nsup & 1 & 4 & 8 & 12 & 16 \\
        \midrule
        0  & 66.2 & 81.9 & 86.4 & 87.9 & 89.1 \\
        5  & 96.2 & 95.7 & 95.8 & 95.9 & 96.0 \\
        10 & 98.3 & 98.2 & 98.2 & 98.2 & 98.4 \\
        15 & 98.8 & 98.8 & 98.8 & 98.7 & 98.8 \\
        19 & 55.3 & 62.3 & 62.1 & 62.3 & 62.1 \\
        \bottomrule
    \end{tabular}
    \caption{Probe accuracy (\%) by denoising step $t$ and completed supervision
steps, without resetting the recursive state between columns.}
    \label{tab:internal_state_acc}
\end{table}


\section{Limitations}
\label{app:limitations}

\ours requires the Thinker's tokens to be spatially aligned with the
conditioning. The conditioning is pooled into a fixed token grid, so each token
covers a fixed region of the image, and the Thinker can attach a constraint to a
location only if that correspondence holds across instances.

To test this we render the Sudoku solution on a larger black canvas under a
random translation, or a random translation and scaling
(Tab.~\ref{tab:scaled_sudoku_limitations_diagnosis}). When the conditioning
remains an untransformed puzzle grid, the correspondence between conditioning
tokens and target cells is destroyed, and both puzzle accuracy and cell accuracy
collapse to chance, at $11.1\%$ for nine digits. Raising the token count from
$81$ to $144$ does not help. Applying the same transform to the conditioning
restores the correspondence and recovers partial performance, at $3.4\%$ puzzle
accuracy and $44.3\%$ cell accuracy, so the failure is one of alignment rather
than of task difficulty.

Constrained scenes show the same requirement
(Tab.~\ref{tab:clevr_limitations_diagnosis}). Relation tokens alone give
$51.2\%$ spatial accuracy and $7.8\%$ attribute binding. Revealing a fully
denoised tenth of the target image does not help, nor does supplying a learned
embedding of a reference crop for each attribute combination, added to the
corresponding object token, which improves attribute binding to $21.2\%$ but
leaves spatial accuracy unchanged. Only the centroid mask, which localises each
object in the image, raises both. \ours therefore depends on conditioning that
carries spatial structure, and does not recover that structure from an
unanchored symbolic specification.

\begin{table}[H]
    \centering
    \begin{tabular}{l l c c c}
        \toprule
        Conditioning & Puzzles & TRM num tokens & Puzzle acc. & Cell acc. \\
        \midrule
        puzzle & translated+scaled & 81 & 0.0 & 11.1 \\
        puzzle & translated & 81 & 0.0 & 11.1 \\
        puzzle & translated & 144 & 0.0 & 11.1 \\
        aligned puzzle & translated+scaled & 81 & 3.4 & 44.3 \\
        \bottomrule
    \end{tabular}
    \caption{MNIST Sudoku with the solution rendered on a larger canvas under a
random translation, or translation and scaling. \emph{Aligned} applies the same
transform to the conditioning. Cell accuracy of $11.1\%$ is chance.} \label{tab:scaled_sudoku_limitations_diagnosis}
\end{table}

\begin{table}[H]
    \centering
    \begin{tabular}{l c c}
        \toprule
        Conditioning & Attr. binding (\%) & Spatial acc. (\%) \\
        \midrule
        tokens & 7.8 & 51.2 \\
        tokens + part of solution & 8.3 & 50.4 \\
        tokens + visual attributions & 21.2 & 51.6 \\
        tokens + centroids & 53.8 & 89.4 \\
        centroids with attributes & 54.8 & 91.2 \\
        \bottomrule
    \end{tabular}
    \caption{Constrained scene composition under different conditioning. Only
conditioning that localises objects in the image improves spatial accuracy. The
final row, in which the mask carries one channel per attribute and so states
exactly where each object belongs, is a sanity check rather than a proposed
setting.}

    \label{tab:clevr_limitations_diagnosis}
\end{table}

\section{SRM on \textsc{extreme}}
\label{app:srm_diagnosis}

SRM \citep{wewer2025srm} denoises the grid one cell at a time, in an order
predicted from the model's own uncertainty. Each cell is committed to as it is
generated, and nothing in the sampler can revisit that commitment once made. On
\textsc{hard} this is largely harmless, since most puzzles admit many solutions
and an early choice can usually be completed into one of them. On
\textsc{extreme} it is fatal, as a single solution exists and any departure from
it cannot be repaired.

\begin{figure}[h]
    \centering
    \includegraphics[width=0.6\linewidth]{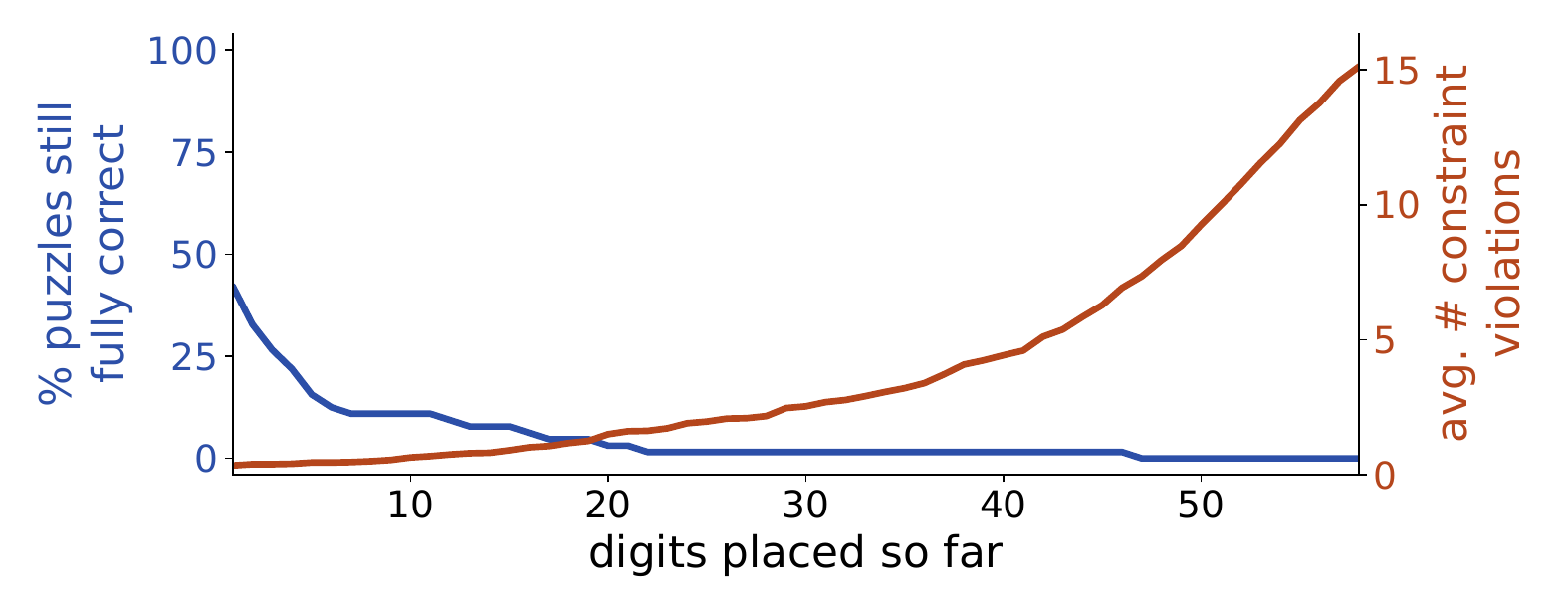}
    \caption{As cells are denoised one by one, the fraction of samples still
    consistent with the unique solution (blue) falls below $50\%$ after the
    first cell, while the number of rules violated on the board (orange) stays
    near zero. SRM's commitments conflict with nothing already present and are
    therefore invisible to it, yet most have already ruled out the only valid
    completion.}
    \label{fig:srm_diagnosis}
\end{figure}

To show that this is what happens, we track two quantities as cells are placed:
the fraction of samples still consistent with the unique solution, and the
number of Sudoku constraints violated on the board so far
(Fig.~\ref{fig:srm_diagnosis}). The two diverge immediately. Violations stay
near zero throughout the early placements, so nothing on the board indicates
that anything has gone wrong, yet consistency with the solution falls below
$50\%$ after the first cell alone. SRM's early commitments are locally
unimpeachable and globally wrong, and because the model can only see the board
it is building, it has no way to detect the difference. This is the same
distinction between locally and globally visible violations that
Sec.~\ref{sec:diagnosis} examines directly.

\section{Efficient allocation of the recursion budget}
\label{app:efficiency}

Three refinements reduce the compute \ours needs without loss of accuracy.
Together they produce the efficient variant of
Fig.~\ref{fig:flops_scaling_results}.

\begin{figure}[H]
    \centering
    \includegraphics[width=0.42\linewidth]{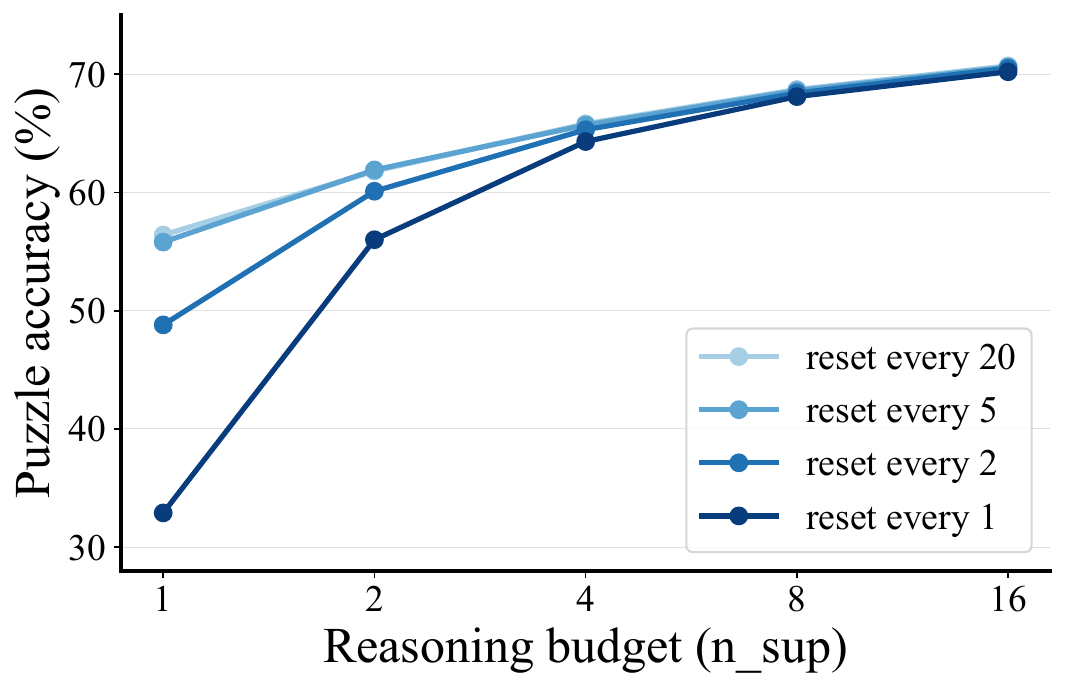}
        \caption{Puzzle accuracy against reasoning budget $n_{\text{sup}}$ for
        different reset intervals. The Thinker's state carries across denoising steps
        , improving accuracy by $23$ points at low budgets.}
    \label{fig:reset_internal_states}
\end{figure}

\paragraph{The reasoning state persists across denoising steps.} If the Thinker
accumulates knowledge about the puzzle, its state should remain useful from one
denoising step to the next rather than needing to be rebuilt. We test this by
varying how often the state is reset (Fig.~\ref{fig:reset_internal_states}).
Resetting at every step costs a great deal at small budgets, giving $32.8\%$
accuracy at a single recursion against $56.1\%$ when the state is carried for
twenty, and the gap closes only once the budget is large enough to rebuild the
state from scratch each time, with all settings converging near $70\%$ at
sixteen recursions. What the Thinker computes is thus worth preserving, which is
what a reasoning state should be.

\begin{figure}[H]
    \centering
    \includegraphics[width=0.42\linewidth]{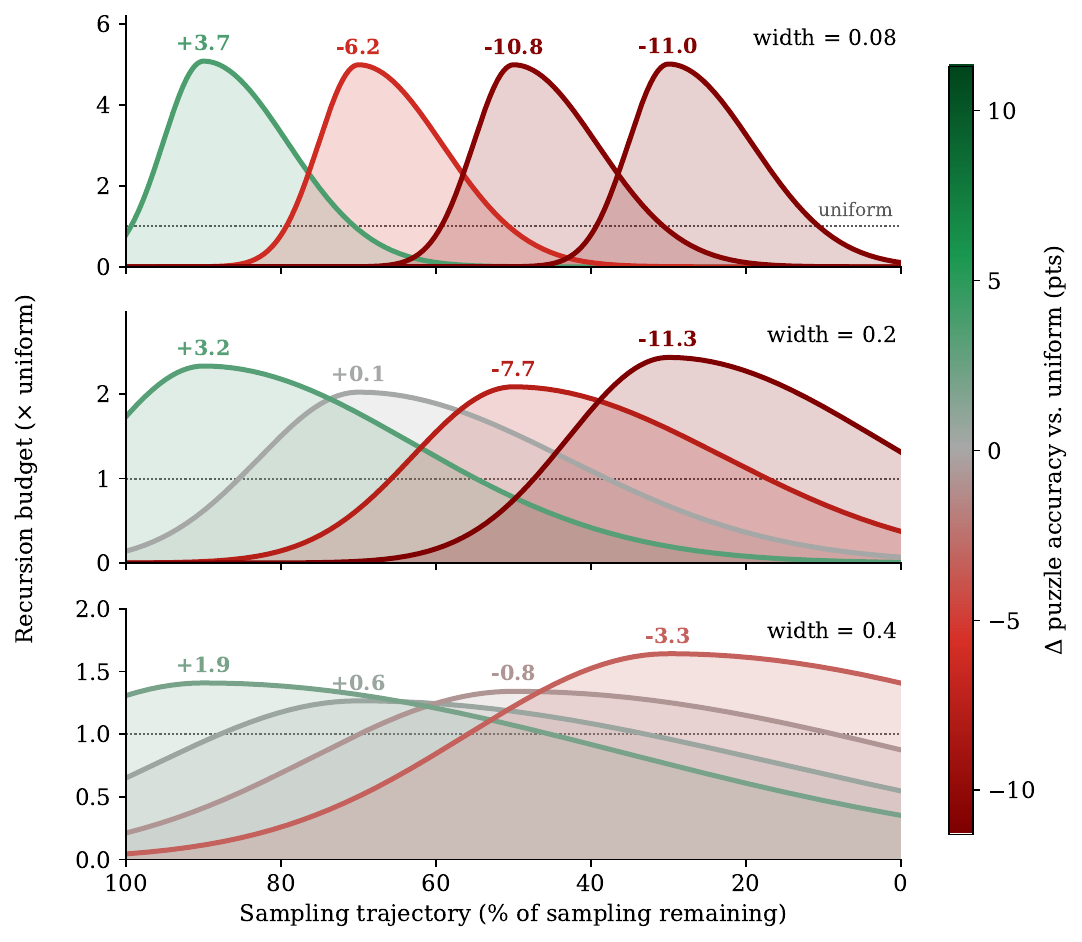}
        \caption{Change in puzzle accuracy when a fixed recursion budget is
        concentrated at different points of the trajectory (columns) with different
        spread (rows), relative to uniform allocation. Allocating early helps and
        allocating late costs up to $11.3$ points.}
        \label{fig:budget_allocation_ablation}
\end{figure}

\begin{figure}[H]
    \centering
    \includegraphics[width=0.42\linewidth]{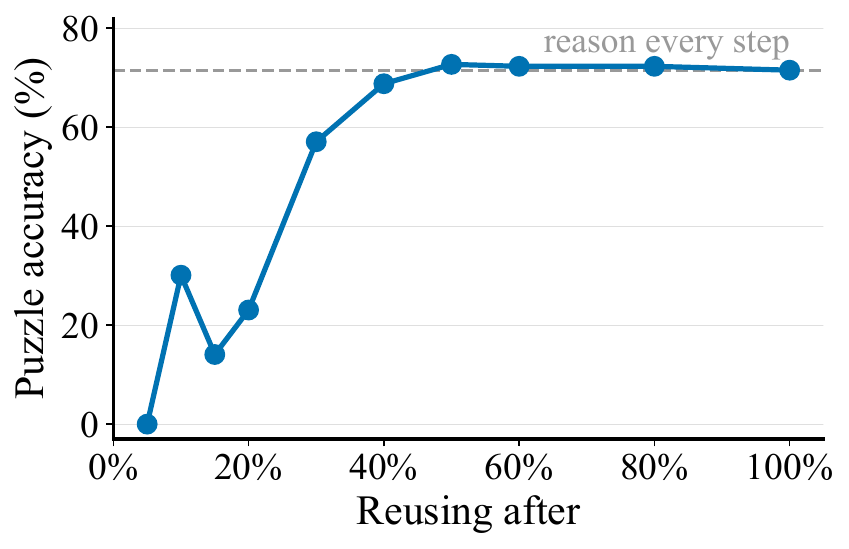}
        \caption{Puzzle accuracy when the Thinker stops reasoning after a given
        fraction of sampling and its last output is reused thereafter. Reasoning for
        the first half suffices, slightly exceeding reasoning at every step
        ($72.5\%$ against $71.2\%$) at half the cost.}
        \label{fig:reuse_steering}
\end{figure}

\paragraph{Reasoning is only needed early.} The Thinker performs the same
computation at every denoising step, though the diagnosis of
Sec.~\ref{sec:diagnosis} suggests this is wasteful, since most of the rule is
resolved in the first few steps. We first redistribute a fixed recursion budget
across the trajectory, concentrating it at different points
(Fig.~\ref{fig:budget_allocation_ablation}). Allocating early helps and
allocating late hurts, by up to $11.3$ points when the budget is centred at
$70\%$ of sampling. We then remove reasoning entirely after a chosen step,
reusing the Thinker's last output for the remainder
(Fig.~\ref{fig:reuse_steering}). Reasoning for only the first half of the
trajectory matches reasoning throughout, and reasoning for the first $50\%$
slightly exceeds it, at $72.5\%$ against $71.2\%$, at half the cost.

\section{Full resolution MNIST Sudoku results}
\label{app:full_resolution_sudoku}
In Tab.~\ref{tab:full_res_sudoku} we present the results of \ours on full-resolution MNIST Sudoku datasets. The performance on \textsc{hard} is nearly identical, and while marginally worse on \textsc{extreme}, it is still significantly better than all the competing methods.

\begin{table}[H]
    \centering
    \begin{tabular}{l c c}
        \toprule
        Resolution & \textsc{hard} & \textsc{extreme} \\
        \midrule
        144x144 & 0.925 & 0.712 \\
        252x252 (full-resolution) & 0.926 & 0.682\\
        \bottomrule
    \end{tabular}
    \caption{Puzzle accuracy comparison of \ours on different resolutions of MNIST Sudoku.}

    \label{tab:full_res_sudoku}
\end{table}

\section{Exploration traces in fine-tuned Bagel}
\label{app:bagel_traces}
\begin{figure}[h]
    \centering
    \includegraphics[width=\linewidth]{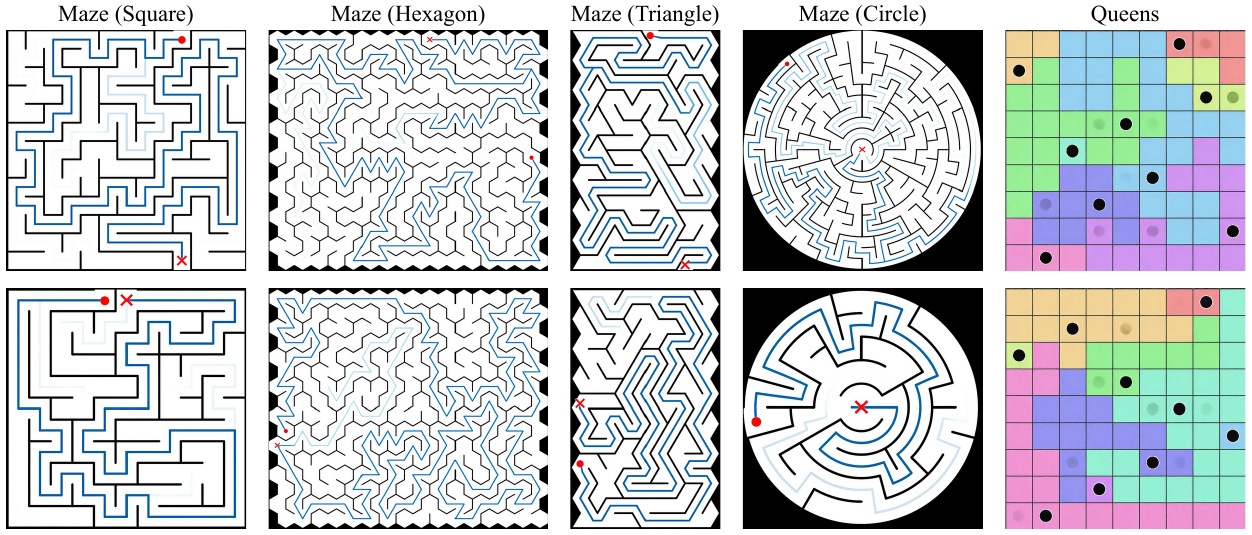}
    \caption{Examples of solutions to mazes and queens puzzles generated by Bagel.}
    \label{fig:bagel_traces}
\end{figure}

On mazes, fine-tuned Bagel reaches near-complete \textsc{coverage} ($99.99\%$)
yet a \textsc{violation} of $12.8\%$, and only $43.4\%$ of its samples are
exactly correct. Fig.~\ref{fig:bagel_traces} shows why. In most samples, the
correct path is drawn in full, but it is surrounded by semi-transparent traces
of other paths. We interpret these as paths the model considered and
rejected but could not fully erase, since the image is the only state it has.
\ours evaluates candidates in the Thinker's latent state, and its samples show
no such traces.

